\documentclass[letterpaper]{article} 
\usepackage{aaai2026}  
\usepackage{times}  
\usepackage{helvet}  
\usepackage{courier}  
\usepackage[hyphens]{url}  
\usepackage{graphicx} 
\usepackage{natbib}  
\usepackage{caption} 
\usepackage{algorithm}
\usepackage{algorithmic}
\usepackage{makecell}

\usepackage{newfloat}
\usepackage{listings}
\DeclareCaptionStyle{ruled}{labelfont=normalfont,labelsep=colon,strut=off} 
\floatstyle{ruled}
\newfloat{listing}{tb}{lst}{}
\floatname{listing}{Listing}
\usepackage{amsmath}
\usepackage{amssymb}
\usepackage{booktabs}
\usepackage{multirow}
\usepackage{subcaption}
\usepackage{tabularx}
\usepackage[most]{tcolorbox}   
\usepackage{xcolor} 
\nocopyright 

\title{ReInAgent: A Context-Aware GUI Agent Enabling Human-in-the-Loop Mobile Task Navigation}
\author {
    Haitao Jia\textsuperscript{\rm 1},
    Ming He\textsuperscript{\rm 2},
    Zimo Yin\textsuperscript{\rm 2},
    Likang Wu\textsuperscript{\rm 3},
    Jianping Fan\textsuperscript{\rm 2},
    Jitao Sang\textsuperscript{\rm 1}
}
\affiliations {
    \textsuperscript{\rm 1}Beijing Jiaotong University, 
    \textsuperscript{\rm 2}AILab at Lenovo Research, 
    \textsuperscript{\rm 3}Tianjin University\\
    23120356@bjtu.edu.cn, heming01@foxmail.com, zmyin20@fudan.edu.cn \\
    wulk@tju.edu.cn, jfan1@lenovo.com, jtsang@bjtu.edu.cn
     
}

\begin{document}

\maketitle

\begin{abstract}
Mobile GUI agents exhibit substantial potential to facilitate and automate the execution of user tasks on mobile phones. However, exist mobile GUI agents predominantly privilege autonomous operation and neglect the necessity of active user engagement during task execution. This omission undermines their adaptability to information dilemmas including ambiguous, dynamically evolving, and conflicting task scenarios, leading to execution outcomes that deviate from users’ genuine requirements and preferences. To address these shortcomings, we propose ReInAgent, a context-aware multi-agent framework that leverages dynamic information management to enable human-in-the-loop mobile task navigation. ReInAgent integrates three specialized agents around a shared memory module: an information-managing agent for slot-based information management and proactive interaction with the user, a decision-making agent for conflict-aware planning, and a reflecting agent for task reflection and information consistency validation. Through continuous contextual information analysis and sustained user-agent collaboration, ReInAgent overcomes existing approaches’ reliance on clear, static task assumptions. Consequently, it enables more adaptive and reliable mobile task navigation in complex, real-world scenarios. Experimental results demonstrate that ReInAgent effectively resolves information dilemmas and produces outcomes that are more closely aligned with users’ genuine preferences. Notably, on complex tasks involving information dilemmas, ReInAgent achieves a 25\% higher success rate than Mobile-Agent-v2.
\end{abstract}


\section{Introduction}

Task automation agents significantly enhance productivity by autonomously performing routine tasks, allowing humans to focus on complex activities. Recent breakthroughs in Large Language Models (LLMs) and Multimodal Large Language Models (MLLMs) \cite{hurst2024gpt,palm,touvron2023llama,Chain-of-thought,react,Tree-of-thoughts,Tool-learning-with-foundation-models,toolformer} have endowed AI agents with robust reasoning and planning abilities, empowering them to utilize tools autonomously for sophisticated task execution \cite{bubeck2023sparks,mirchandani2023large,liang2023code,wang2023describe,yao2023react,zhao2024expel,talebirad2023multi}. 

\begin{figure}[t]
    \centering
    \includegraphics[width=\linewidth]{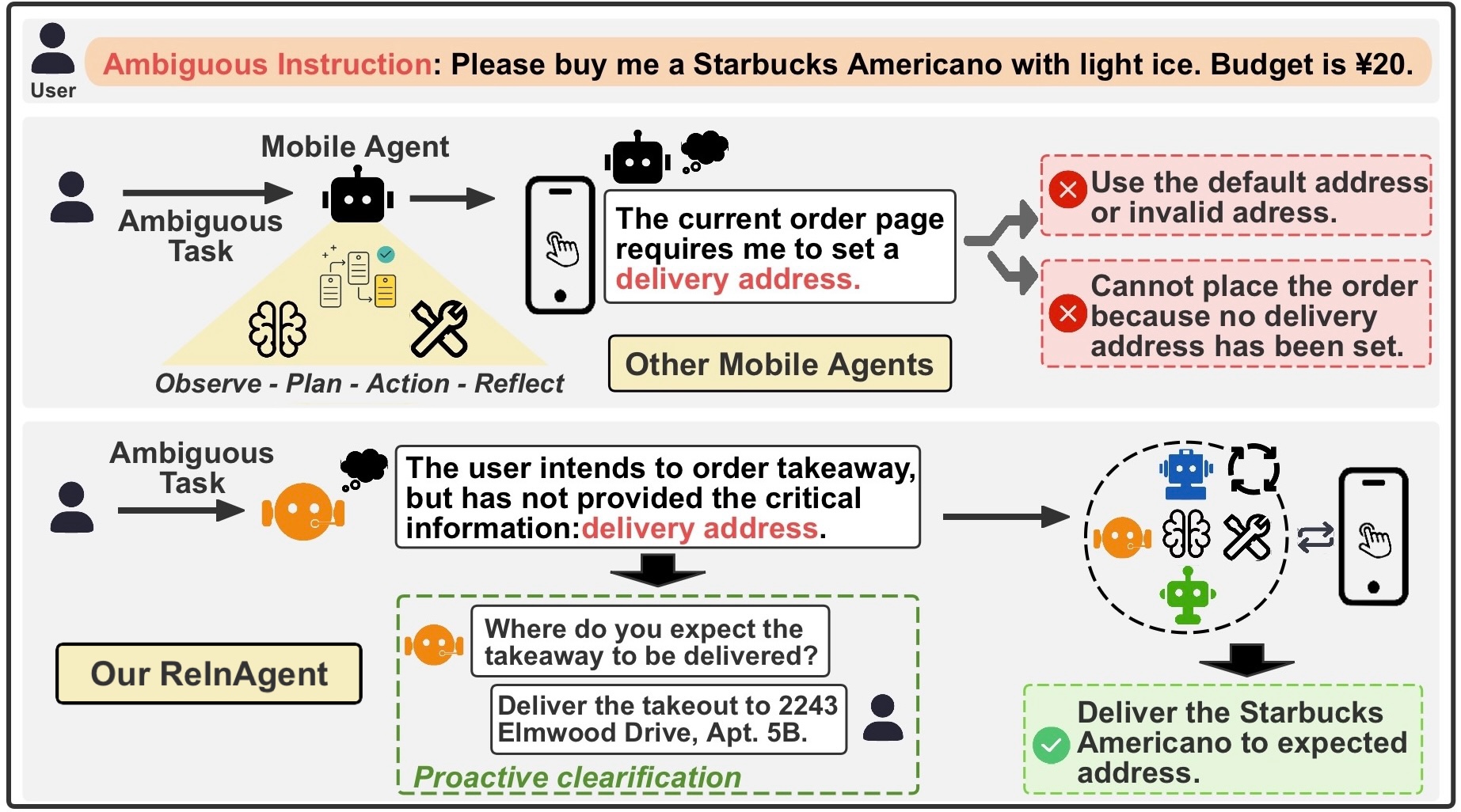}
    \caption{Illustration of the comparison between our framework and other mobile agents when encountering ambiguous task.}
    \label{fig:comparison framework}
\end{figure}

Mobile phones, as a controlled environment with rich interaction scenarios, have become prominent platforms for GUI-based agent research, particularly in navigation and task automation \cite{wang2024gui,wu2024foundations}. Current mobile GUI agents predominantly rely on the assumption that user instructions are both complete and unambiguous from the outset. This simplified assumption fails to capture the inherent uncertainties and variabilities present in real-world settings. Moreover, they typically pursue fully autonomous task execution without integrating user-agent collaboration. For example, in the task of ordering a Starbucks Americano shown in Figure~\ref{fig:comparison framework}, user instructions often omit critical information such as the delivery address, which significantly impairs the agent’s ability to select the correct store and configure the order properly. Furthermore, when the item price exceeds the user’s budget or additional preferences emerge, a non-interactive agent may enter ineffective loops or make substitutions that deviate from the user’s intent. These issues highlight the necessity of user-agent collaboration to resolve information dilemmas encountered during task execution. Specifically, agents commonly face three critical information dilemmas: (1) ambiguous initial instruction that omits critical details; (2) incremental information supplementation that emerges during task progression; (3) conflicting information that contradicts specified requirements.

Recent studies adopt diverse methods to tackle the challenges posed by these information dilemmas. Qian \textit{et al.} \cite{tellmemore} uses a data-driven method to endow the agent with the ability to predict implicit user intentions from ambiguous instructions before task execution. Zhang \textit{et al.} \cite{askbeforeplan} introduces a Clarification-Execution-Planning framework to complete the instruction clarification before task execution. Liu e\textit{t al.} \cite{autoglm} incorporates proactive user-agent interactions during execution to address the incremental information requirements. Nevertheless, these methods fail to resolve all three information dilemmas simultaneously within a single task cycle and fall short of enabling dynamic task evolution during execution.

In this paper, we propose ReInAgent, a human-in-the-loop multi-agent framework designed to resolve information dilemmas and enable dynamic task evolution for task automation on mobile phones. ReInAgent addresses the three information dilemmas using a dynamic task-slot management mechanism combined with proactive user-agent interaction, inspired by traditional dialogue management systems. To be specific, ReInAgent comprises three specialized agents, namely the Information-managing Agent (ImA), the Decision-making Agent (DmA) and the Reflecting Agent (RA). They collaborate through a shared memory module, enabling information exchange and coordinated decision-making and reflection. Acting as the interface for user-agent interaction, the ImA clarifies ambiguities in initial task instructions and proactively engages with the user during execution to supplement incremental information and resolve conflicts. To initiate task execution, the DmA first decomposes the clarified task instruction into executable subtasks, followed by iterative execution. In each iteration, it interprets the current screen state, makes decisions, and operates the mobile device using tools. When conflicting information arises, it can seek assistance from the ImA to solicit user feedback, ensuring decisions remain aligned with user intent. Notably, each resolution of an information dilemma involves updating the task context, including task-slot values, subtasks, etc. After each action execution, the RA reflects the execution result, information consistency, and task progression to guide the next round of task iteration. Moreover, RA has a history summarization mechanism to cut overhead in lengthy task contexts.

Our contributions can be summarized as follows:
\begin{itemize}
    \item We propose ReInAgent, a multi-agent collaborative framework that integrates proactive user-agent interaction to autonomously operate mobile devices and accomplish complex real-world tasks.
    \item We introduce a slot-based information management mechanism to resolve information dilemmas and enable dynamic task evolution during execution. The ImA dynamically generates and updates task slots through proactive interaction with the user, thereby guiding the DmA’s execution. Concurrently, the RA evaluates information consistency, providing iterative feedback to support task evolution.
    \item We demonstrate the effectiveness of ReInAgent on real-world mobile tasks through detailed task-level and information-level evaluations, highlighting notable improvements in dynamic information management and execution performance. Furthermore, we empirically show that integrating app-specific knowledge significantly enhances its automation capabilities.
    
\end{itemize}

\section{Related Works}

\subsection{Mobile GUI Agent}
In recent years, breakthroughs in LLMs and MLLMs have accelerated the rapid development of AI Agent research based on multimodal large models \cite{palm,touvron2023llama,Chain-of-thought,react,Tree-of-thoughts,Tool-learning-with-foundation-models,toolformer}. Given that mobile devices offer controllable operational environments with rich task scenarios, mobile GUI agents emerge as a critical research focus. Yan \textit{et al.} \cite{yan2023gptinwonder} uses GPT-4V to understand the screenshots with annotations and make decisions to operate the phone. Zhang \textit{et al.} \cite{appagent} designs an agent that automatically explores the app and applies the knowledge learned to task execution. Wang et al. \cite{mobile-agent} utilizes a visual perception module to eliminate the reliance on the application’s XML files. Wang \textit{et al.} \cite {Mobile-agent-v2} further designs a multi-agent collaboration framework with a memory module to store the focus content and external knowledge to assist task execution.
 Hong \textit{et al.} \cite{cogagent} adopts data-driven approaches to improve the visual understanding and localization abilities of the mobile agent. 
 Liu \textit{et al.} \cite{autoglm} explores the self-evolution of the mobile agent via reinforcement learning during the task.

\subsection{Human-agent Interaction}
Recent research increasingly emphasizes human-agent collaborations to enhance agent capabilities \cite{deng2024multi}. For example, Qian \textit{et al.} \cite{tellmemore} fine-tunes LLMs to address ambiguities in user instructions via multi-round dialogue with the user before executing the tasks. In contrast, Lu \textit{et al.} \cite{proactiveAgent} emphasizes agent proactivity by enabling it to anticipate user intent and proactively deliver relevant information to the user in scenarios without specified task instructions. Zhang \textit{et al.} \cite{askbeforeplan} designs a framework that not only engages in communication with the user but also interacts with external tools and web environments to gather information to clarify task details. However, these works still fall short in resolving the three information dilemmas and neglect the evolution of user tasks during a task cycle in complex dynamic environments. While Liu \textit{et al.} \cite{autoglm} leverages reinforcement learning to enhance the agent’s ability to supplement information via user-agent interactions during the decision-making process, it does not take into account the issue of conflicting information. Our work introduces a human-in-the-loop multi-agent framework with a slot-based information management mechanism to effectively address the information dilemmas and realize the dynamic evolution of user tasks.

\begin{figure*}[t]
    \centering
    \includegraphics[width=\linewidth]{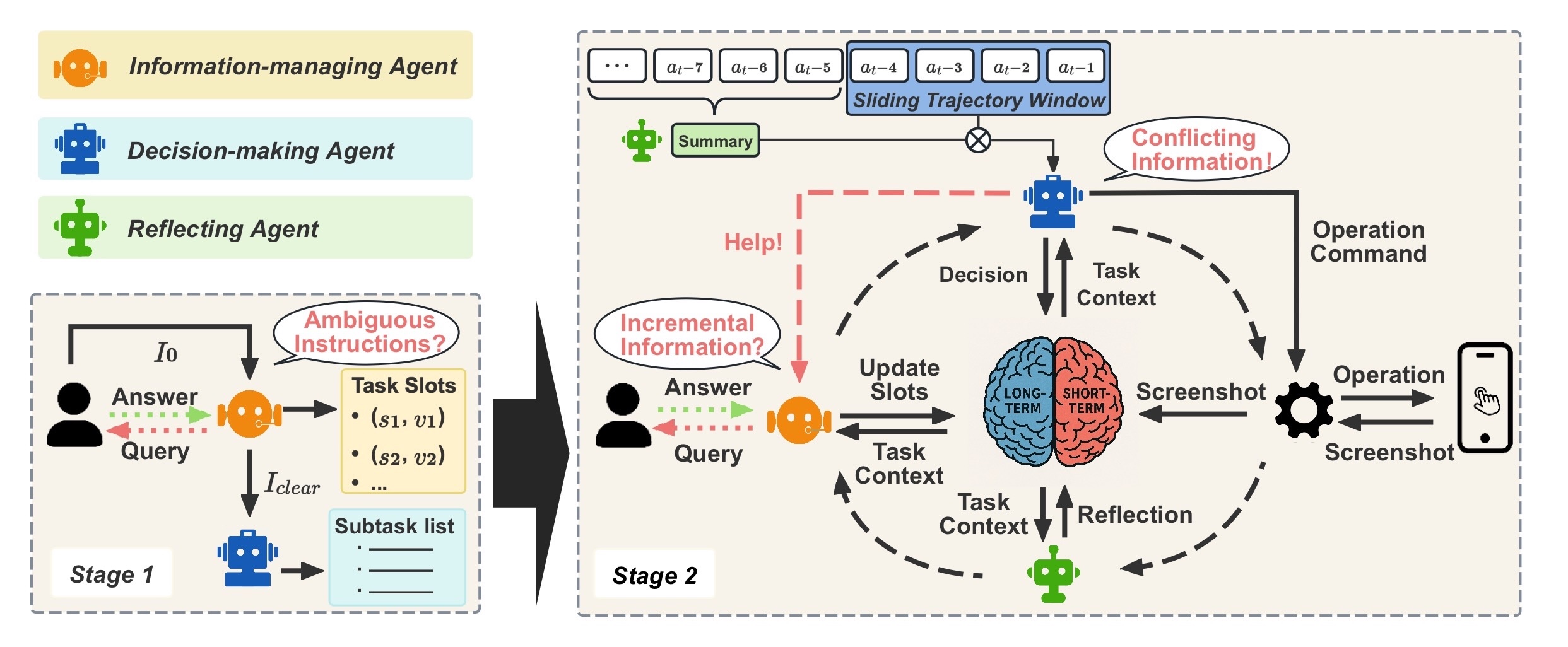}
    \caption{Overview of the framework of ReInAgent. The red and green dashed lines indicate the interactions among the DmA, the ImA, and the user under different information dilemmas.}
    \label{fig:workflow}
\end{figure*}

\section{ReInAgent}
\label{headings}
ReInAgent is a human-in-the-loop multi-agent framework that autonomously operates the mobile phone to accomplish complex tasks. Moreover, the framework incorporates a dynamic information management mechanism via proactive interaction with the user, significantly enhancing its flexibility and effectiveness in task automation.

\subsection{Framework}
As illustrated in Figure~\ref{fig:workflow}, ReInAgent consists of three specialized agents along with Memory and Tool modules:
\begin{equation}
\text{ReInAgent} = \langle \text{Agents}, M, U, T, \Omega \rangle,
\end{equation}
where \(\text{Agents} = \{\text{ImA}, \text{DmA}, \text{RA}\}\); \(M\) denotes the memory module; \(U\) represents the user; \(T\) is the localization tool; and \(\Omega\) defines the collaboration rules among the three agents, the memory module, the tools and the user. ReInAgent adopts a two-stage automation framework, consisting of a task pre-processing phase followed by iterative task execution.

\subsubsection{Task Pre-processing.} In order to mitigate the ambiguity in the initial user instruction \(I_0\), the ImA performs ambiguity analysis and instruction clarification at this stage. 
This process is detailed further in the Info-managing Agent section. Following this, the DmA decomposes the clarified instruction \(I_{\text{clear}}\) into a detailed subtask list \(L_{\text{subtask}} = \{l_1, l_2, ..., l_m\}\), which is shown in the Decision-making Agent section.

\subsubsection{Iterative Task Execution.} After the task pre-processing, three specialized agents collaborate to execute the user's task iteratively. Each iteration \(t\) comprises four core steps: information management, decision-making, executing actions, and reflecting. 
Specifically, the ImA first conducts the incremental information analysis. Subsequently, the DmA observes the screen state and makes decisions to operate the mobile phone with the tools. Notably, the DmA can seek assistance from the ImA if it detects the information conflicts during decision-making. Finally, the RA reflects on actions and task progression and validates information consistency to guide the next iteration.
\subsection{Information-managing Agent} \label{Info-managing Agent}
Drawing on traditional dialogue management systems, the ImA serves as the interface for user-agent interaction and employs a slot-based mechanism to manage task-related information efficiently during each iteration. Specifically, it is responsible for instruction clarification, incremental information supplementation and information conflict resolution.
\subsubsection{Instruction Clarification.} Based on the initial instruction, the ImA first infers the user’s task intent and then generates a set of task-relevant slots \(S = \{s_1, s_2, \dots, s_n\}\). This process is formalized as:
\begin{equation}
    \begin{gathered}
     \hat{z} = \arg\max_{z \in \mathcal{Z}} \text{ImA}_{\text{intent}}(I_0),\\
        S = \text{ImA}_{\text{slot}}(\hat{z}, I_0) = \{(k_1, v_1),  ..., (k_n, v_n)\},
    \end{gathered}
\end{equation}
where \(\text{ImA}_{\text{intent}}\) maps the \(I_0\) to an intent \(\hat{z}\) from the intent space \(\mathcal{Z}\). \(\text{ImA}_{\text{slot}}\) generates \(S\) based on the predicted \(\hat{z}\) and the \(I_0\). Each slot \(s_i = (k_i, v_i)\) comprises a key \(k_i\), which denotes a required information category, and a value \(v_i\), which is either extracted from the \(I_0\) or obtained by querying the user. Collectively, these slots organize the essential information required to execute the user’s intended task.
After that, the ImA transforms the ambiguous \(I_0\) into \(I_{\text{clear}}\) through:
\begin{equation}
I_{\text{clear}} = \text{ImA}_{\text{clear}}(I_0, S),
\end{equation}
where \(\text{ImA}_{\text{clear}}\) integrates the \(S\) with \(I_0\) to produce \(I_{\text{clear}}\).

\begin{figure}[!t]
    \centering
    \begin{subfigure}{\linewidth}
        \centering
        \includegraphics[width=\linewidth]{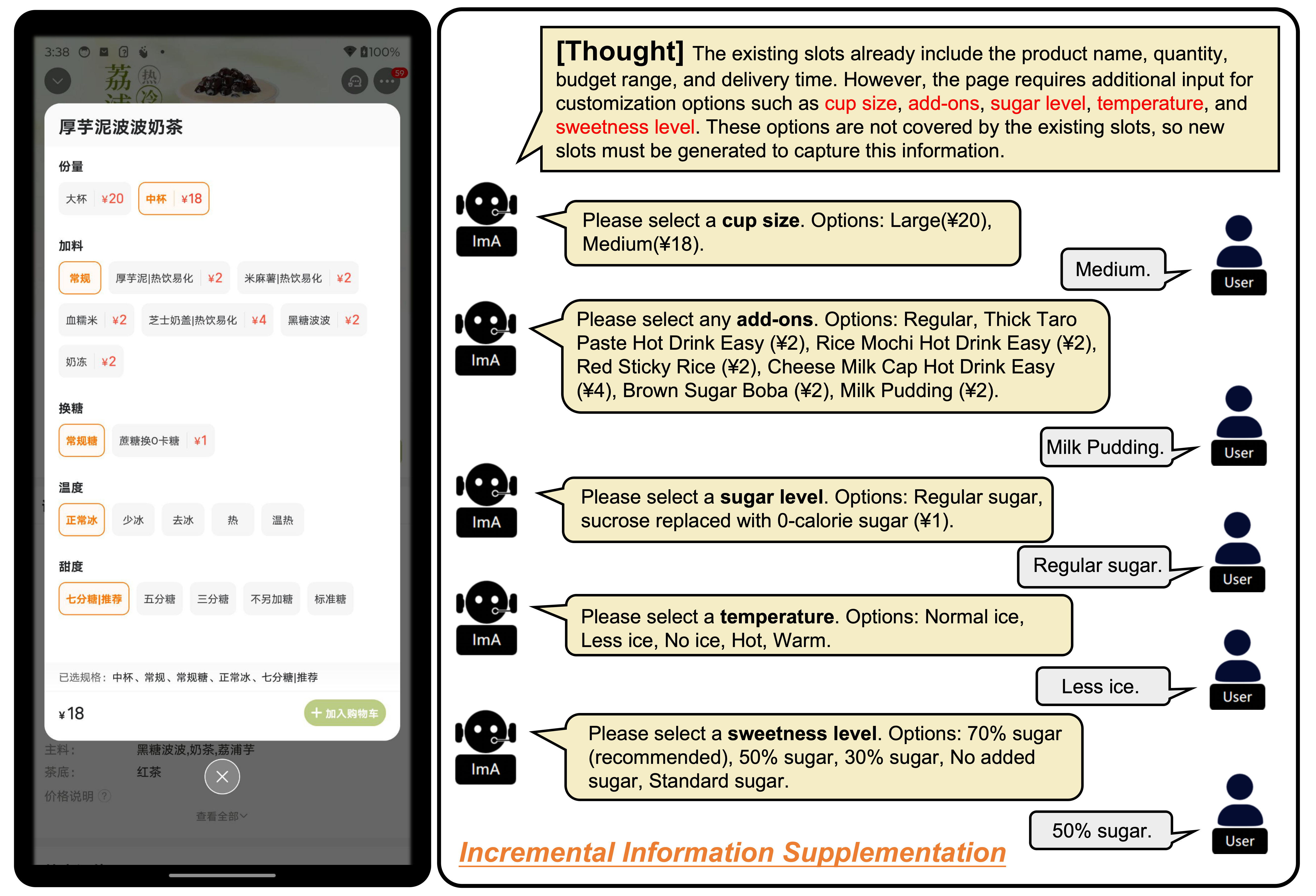}
        \caption{Task: please help me order a thick taro boba milk tea on the Meituan app.}
        \label{subfig:info_more}
    \end{subfigure}
    \hspace{2mm}
    \begin{subfigure}{\linewidth}
        \centering
        \includegraphics[width=\linewidth]{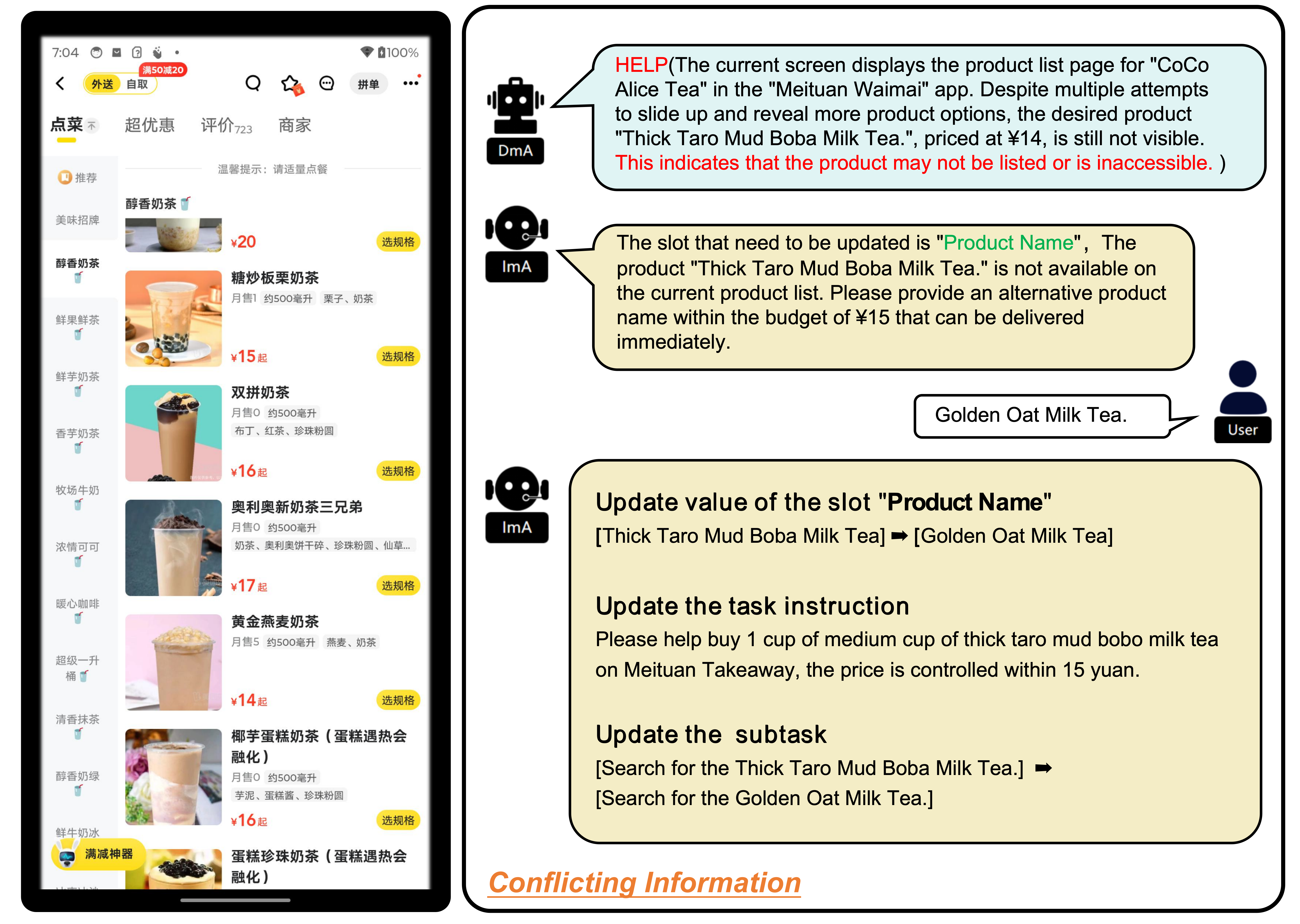}
        \caption{Task: help me order a thick taro boba milk tea with under 14 RMB.}
        \label{subfig:info_con}
    \end{subfigure}
    \caption{Illustration of two types of information dilemmas encountered during the iterative task execution stage. Figure (a) shows incremental information supplementation, while Figure (b) illustrates conflicting information .}
    \label{fig:info_dilemma}
\end{figure}

\subsubsection{Incremental Information Supplementation.} As shown in Figure~\ref{subfig:info_more}, certain screen states demand incremental information to perform selections. Upon detecting a need for additional information, the ImA generates the required slots. This process is formalized as:
\begin{equation}
S_t^+ = S_t \cup \text{ImA}_{\text{add}}(O_t, S_t, l_t),
\end{equation}
where \(\text{ImA}_{\text{add}}\) generates additional slots by analyzing the observation \(O_t\), current slots \(S_t\), and the current subtask \(l_t\), and \(S_t^+\) denotes the updated slot set after supplementation. Subsequently, the ImA integrates the supplemented slots into the memory module, updating both the task instruction and the corresponding subtasks.

\subsubsection{Information Update.} During task execution, the screen states may contain conflicting information, such as unavailable products in a specified store, as illustrated in Figure~\ref{subfig:info_con}. In such situation, the ImA analyzes the dilemma identified by the DmA, determines the necessary slot updates, and proactively notifies the user of any conflicts while requesting updated slot values. Given a detected inconsistency:
\begin{equation}
\Delta_t = \{(s_i, o_i) | s_i \in S_t, o_i \in O_t, \mathbb{I}(s_i, o_i) = 0\},
\end{equation}
where \(\mathbb{I}\) is a binary function that evaluates consistency between slot information \(s_i\) and observed screen content \(o_i\). The ImA updates the slot set by removing inconsistent slots and adding new slots with corrected values. This slot update process follows:
\begin{equation}
    \begin{gathered}
        \tilde S_t 
  = S_t \setminus \{\,s_i \mid (s_i,o_i)\in\Delta_t\}, \\
  S_{t+1}
  = \tilde S_t \;\cup\;\{(k_i,v_i^*) \mid (s_i,o_i)\in\Delta_t\},
    \end{gathered}
\end{equation}
where \(v_i^*\) represents updated slot values elicited through user interaction. Upon receiving the user's feedback, the ImA first discards the outdated slot entries in \(S_t\) to form \(\tilde S_t\), and then merges the updated pairs \(\{(k_i,v_i^*)\}\) into \(\tilde S_t\), yielding the dynamically evolved state \(S_{t+1}\).

\subsection{Decision-making Agent}\label{subsec:dma}
The DmA is responsible for hierarchical planning and sequential decision-making within our framework.

\subsubsection{Hierarchical Planning.}
Decomposing a complex task into multiple simple subtasks enhances the feasibility of task automation \cite{huang2024understanding}. At the task pre-processing stage, the DmA decomposes the user's task into executable subtasks based on \(I_{\text{clear}}\) and injected external knowledge \(K\):
\begin{equation}
L_{\text{subtask}} = \text{DmA}_{\text{dec}}\left(I_{\text{clear}}, K\right),
\end{equation}
where \(\text{DmA}_{\text{dec}}\) represents the task decomposition strategy. This decomposition can be formulated as an optimization problem:
\begin{equation}
L_{\text{subtask}} = \arg\min_{L} \sum_{i=1}^{|L|} \Phi(l_i) + \alpha \cdot \mathcal{D}(L, I_{\text{clear}}, S, K),
\end{equation}
where \(\Phi(l_i)\) represents the execution complexity of subtask \(l_i\), \(\mathcal{D}\) measures semantic distance between \(L_{\text{subtask}}\) and the task requirements, and \(\alpha\) controls trade-off between complexity and completeness.

\subsubsection{Sequential Decision-making.} During each iteration \(t\), DmA analyzes the \(O_t\) and the task context to select an appropriate action from the adaptive action space \(\mathcal{A}\), with details available in Appendix. Notably, an excessively long task‑context history can cause LLMs to overlook critical historical information during decision-making \cite{liu2024lost} and lead to hallucination \cite{qiu2024longhalqa}. To mitigate the overhead of long histories, DmA employs a sliding trajectory window that attends only to the most recent \(N\) actions. Experimental results in Appendix confirm that ReInAgent performs best on the task when the trajectory window size is set to \(4\). Simultaneously, earlier steps are abstracted into a summary of completed subtasks and insights acquired by the RA. The decision-making process can be formalized using probabilistic action selection:
\begin{equation}
\begin{gathered}
  C_t = \bigl(I_{\mathrm{clear}},\,L_{\mathrm{subtask}}, S,\,K,\,H_t,\,T_a^{(t-4:t-1)} \bigr),\\
  A_t = \arg\max_{a\in\mathcal{A}} P\bigl(a \mid O_t,\,R_{t-1},\,C_t\bigr),
\end{gathered}
\end{equation}
where \(C_t\) represents the task context, \(H_t\) summarizes the action history, \(R_{t-1}\) refers to the RA’s previous reflection result, and \(T_a^{\left(t-4:t-1\right)}\) denotes the trajectory of the four most recent actions. With a decision made, the DmA utilizes localization and ADB tools to precisely operate the phone. 

\subsection{Reflecting Agent}
The explicit reflection mechanism is the key driver of better reasoning, lower hallucination, and higher overall task success for LLMs and LLM‑powered agents \cite{madaan2023self, shinn2023reflexion, yao2023react}. Consequently, we implement this mechanism within the RA, which also summarizes and abstracts the task history.

\subsubsection{Reflection.} After each action execution, the RA evaluates the effectiveness of \(a_{t-1}\) and verifies alignment between the current screen state and the task slots. Based on this assessment, it determines whether the task has been completed. The reflection process is formalized as:
\begin{equation}
        R_t = \text{RA}_{\text{ref}}\left(a_{t-1}, O_{t-1}, O_{t}, C_t \right),
\end{equation}
where \(R_t\) includes a multi-dimensional assessment:
\begin{equation}
\begin{gathered}
E_t = \text{sim}(O_t, p_{t-1}), \\
Q_t = \text{cons}(S, O_t), \\
P_t = \frac{|\{l_i \in L_{\text{subtask}} \mid \text{completed}(l_i, O_t)\}|}{|L_{\text{subtask}}|},
\end{gathered}
\end{equation}
where \(E_t\) represents action effectiveness measured by the similarity between \(O_t\) and the expected goal \(p_{t-1}\), \(Q_t\) evaluates information consistency between slot values and \(O_t\), and \(P_t\) quantifies task progress as the ratio of completed subtasks to total subtasks.

\subsubsection{Trajectory Summarization.} As described above, we design the trajectory summarizing mechanism to mitigate the overhead from long task contexts. As depicted in Figure~\ref{fig:workflow}, the RA summarizes the action trajectory into completed subtasks and experience after every four actions. This process is defined as:
\begin{equation}
H_t = \text{RA}_\text{sum}\left(H_{t-4}, T_a^{t-4:t-1}\right), \quad t\mod 4 = 0,
\end{equation}
where \(\text{RA}_\text{sum}\) denotes RA's rule of summarization, and \(H_{t-4}\) is the previous history summary. This process can be formulated as a projection operation that preserves task-critical information, subject to the dimensionality constraint \(\text{dim}(h) \leq d_{\text{max}}\):
\begin{equation}
\text{RA}_\text{sum}(H, T) = \arg\min_{h \in \mathcal{H}} \mathcal{L}(h, H \oplus T),
\end{equation}
 where \(\oplus\) denotes sequence concatenation, \(\mathcal{L}\) represents information loss, \(\mathcal{H}\) is the set of possible summaries, and \(d_{\text{max}}\) defines the maximum allowable dimension for the summary.

\begin{table*}[t]
    \label{table1}
    \centering
    \renewcommand{\arraystretch}{1.2}
    \setlength{\tabcolsep}{6pt}
    \begin{tabular*}{0.99\linewidth}{lcccccccccc}
        \toprule
        \multirow{2}{*}{\textbf{Method}} & \multicolumn{6}{c}{\textbf{Task-level}} & \multicolumn{4}{c}{\textbf{Information-level}} \\
        \cmidrule(lr){2-7} \cmidrule(lr){8-11}
        & SR & CR & DA & RA & AE & RE & IC & ICR & SA & CCR \\
        \midrule
        AppAgent          & 0.48 & 0.54 & 0.57 & -    & 0.45 &
        0.50    & 0.30 & -    & -    & -    \\
        Mobile-Agent-v2 + Know.  & 0.57 & 0.66 & 0.64 & 0.68 & 0.58 & 0.70    & 0.38 & -    & -    & -    \\
        \midrule
        ReInAgent               & 0.70 & 0.84 & 0.85 & 0.88 & 0.84 & 0.58 & 0.85 & \textbf{0.93} & 0.94 & 0.70 \\
        ReInAgent + Know.          & \textbf{0.82}(\(\uparrow\)0.12) & \textbf{0.91} & \textbf{0.91} & \textbf{0.95} & \textbf{0.87} & \textbf{0.76}(\(\uparrow\)0.18) & \textbf{0.97}(\(\uparrow\)0.12) & 0.92 & \textbf{0.95} & \textbf{0.78} \\
        \bottomrule
    \end{tabular*}
    \caption{Evaluation results on both task-level and information-level metrics, where \textit{Know.} represents injected app-specific operational knowledge. The symbol \(\uparrow\) denotes a significant improvement in ReInAgent’s performance after knowledge injection compared to before.}
\end{table*}

\subsection{Memory Module}
The memory module serves as an essential mediator for agent collaboration and communication within ReInAgent, playing an indispensable role in maintaining information coherence and coordination across the system. It consists of long-term memory and short-term memory.

\subsubsection{Long-term Memory.} Human proficiency in mobile operation typically relies on experience and familiarity with various apps. Similarly, fundamental mobile operational knowledge and app-specific knowledge are critical for ReInAgent’s task execution. The long-term memory stores these two types of knowledge, effectively guiding decision-making and reflection during task execution, which enhances ReInAgent’s navigation capability and efficiency. 

\subsubsection{Short-term Memory.} The short-term memory stores task-relevant information such as task instructions, subtasks, and intermediate parameters like action trajectories and reflections. Additionally, task slots generated by ImA are also stored in the short-term memory module. 

\section{Experiments}
\subsection{Experimental Setup}\label{Experimental Setup}
\subsubsection{Foundation Model.} 
We employ GPT-4o \cite{hurst2024gpt} as the foundational MLLM for all three agents due to its strong capabilities in screen-state comprehension and analytical reasoning.
To achieve more precise UI element localization, we further incorporate OS-Atlas-base-7B \cite{os-atlas} as a dedicated localization tool for the DmA. Detailed model configurations are provided in Appendix.

\subsubsection{Evaluation Details and Tasks.} We compare ReInAgent with Mobile-Agent-v2 \cite{Mobile-agent-v2} and AppAgent \cite{appagent} under identical experimental settings. To further evaluate the impact of injecting app-specific operational knowledge, we compare ReInAgent’s performance before and after the integration of such knowledge. Notably, to ensure fairness in comparative evaluation, we simultaneously inject the same knowledge into Mobile-Agent-v2.
To assess ReInAgent’s effectiveness under realistic conditions, we conduct a real‑time evaluation on Android devices across four representative daily scenarios including takeaway ordering, hotel reservation, flight booking and shopping online. These scenarios are deliberately chosen to reflect common task patterns that involve information dilemmas. Task instructions are detailed in Appendix.

\subsubsection{Metrics.} We categorize evaluation into two dimensions: task‐level and information‐level. Task‐level metrics emphasize the agent’s execution performance, measuring success rate (SR), completion rate (CR), decision accuracy (DA), reflection accuracy (RA), action effectiveness (AE), and relative efficiency (RE). Information‐level metrics focus on the management of task‐related information, evaluating information consistency (IC), information capture rate (ICR), slot accuracy (SA), and conflict capture rate (CCR). Detailed descriptions for each metric are provided in Appendix.

\subsection{Result}

\begin{table*}[t]
\label{tab:table2}
\centering
\small
\renewcommand{\arraystretch}{1.2}
\setlength{\tabcolsep}{2pt}
\begin{tabular*}{0.98\textwidth}{@{\extracolsep{\fill}} ll c c c c c c c c c c @{\extracolsep{\fill}}}
\toprule
\multirow{2}{*}{\textbf{Task}} & \multirow{2}{*}{\textbf{App}} &
\multicolumn{6}{c}{\textbf{Task-level}} &
\multicolumn{4}{c}{\textbf{Information-level}} \\
\cmidrule(lr){3-8} \cmidrule(lr){9-12}
& & SR & CR & DA & RA & AE & RE & IC & ICR & SA & CCR \\
\midrule
Shopping Online & JD.com & \textbf{14/15} & \textbf{0.98} & 0.90 & 0.93 & 0.89 & 0.76 & \textbf{0.99} & 0.93 & \textbf{0.98} & 7/8 \\
Takeaway Ordering & Meituan Takeaway & \textbf{14/15} & 0.95 & 0.92 & 0.94 & \textbf{0.93} & \textbf{0.83} & 0.94 & 0.93 & 0.93 & 5/7 \\
Flight Booking & Ctrip & 10/15 & 0.86 & 0.90 & 0.95 & 0.84 & 0.79 & 0.97 & 0.90 & 0.95 & 5/7 \\
Flight Booking & Skyscanner & 12/15 & 0.91 & \textbf{0.94} & \textbf{0.96} & 0.90 & 0.78 & \textbf{0.99} & 0.89 & 0.95 & 5/6 \\
Hotel Reservation & Ctrip & 11/15 & 0.80 & 0.84 & 0.93 & 0.82 & 0.66 & 0.98 & \textbf{0.95} & 0.95 & 5/7 \\
Hotel Reservation & Booking.com & 13/15 & 0.95 & 0.93 & \textbf{0.96} & 0.87 & 0.76 & 0.94 & 0.92 & 0.96 & 5/6 \\

\bottomrule
\end{tabular*}
\caption{Performance of ReInAgent with app-specific operational knowledge across diverse tasks and apps.}
\end{table*}
\subsubsection{Evaluation on Task-level Metrics.} As Table 1 illustrates, even without injecting app-specific operational knowledge, ReInAgent achieves a task success rate approximately 20\% higher than both AppAgent and Mobile-Agent-v2 when encountering information dilemmas, with completion rates improving by 30\% and 18\%, respectively. Decision accuracy increases by 28\% relative to AppAgent and by 21\% compared with Mobile-Agent-v2. Furthermore, ReInAgent’s reflection success rate surpasses that of Mobile-Agent-v2 by 20\%. Although ReInAgent’s relative efficiency remains below that of Mobile-Agent-v2, which is injected with application operating knowledge, it still outperforms AppAgent. Notably, ReInAgent also shows a significant improvement in action effectiveness. Importantly, when app-specific operational knowledge is incorporated, all task-level metrics for ReInAgent improve. Most notably, the success rate increases by 12\% and relative efficiency by 18\%. The significant performance improvement indicates that integrating app-specific knowledge significantly enhances its task automation capabilities.

\subsubsection{Evaluation on Information-level Metrics.} Table 2 demonstrates that ReInAgent’s dynamic information management yields performance closely aligned with user preferences. It achieves an information consistency score of 0.85, substantially above AppAgent’s 0.30 and Mobile‑Agent‑v2’s 0.38. The information‑capture and conflict‑capture rates of ReInAgent reach 0.93 and 0.70, respectively. Moreover, its accuracy of generated slots scores 0.94. After the injection of app-specific operational knowledge, ReInAgent's consistency increases by 12\% and the conflict-capture rate by 8\%, while information-capture and slot accuracy remain essentially unchanged.

\begin{figure*}[t]
    \centering
    \begin{subfigure}[b]{0.8\linewidth}
        \centering
        \includegraphics[width=\linewidth]{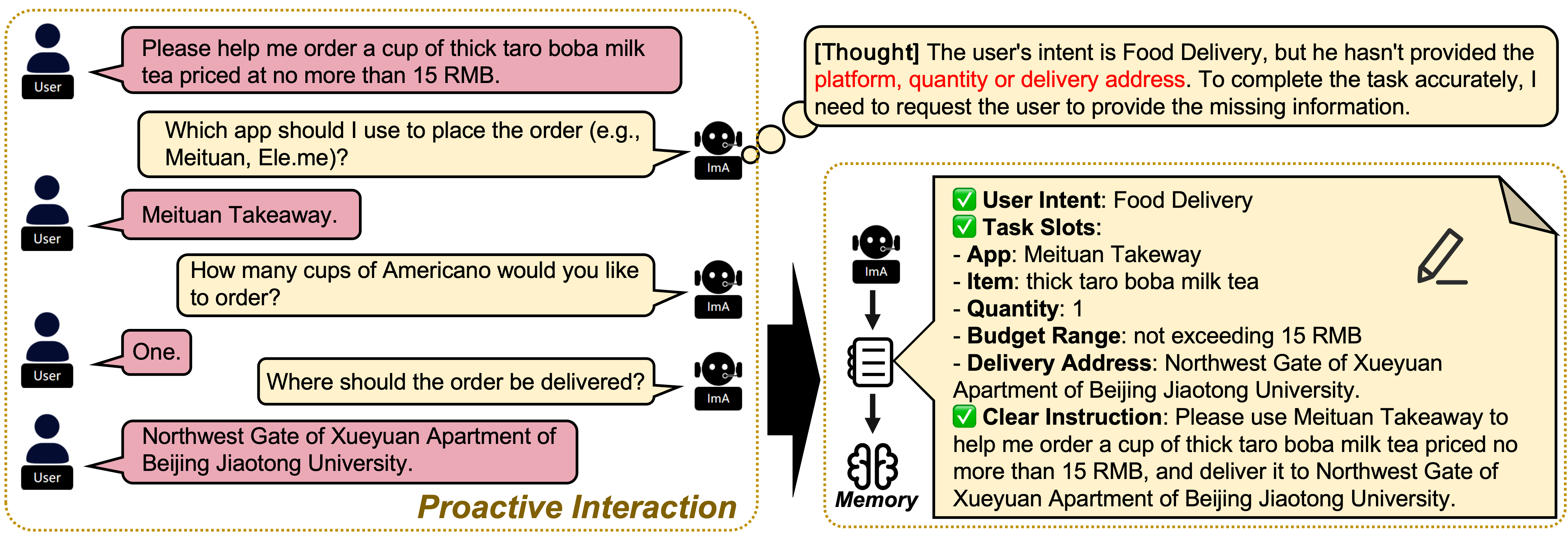}
        \caption{Stage 1: task pre-processing.}
        \label{casestudy:subfig1}
    \end{subfigure}

    \vspace{0.1cm}  

    \begin{subfigure}[b]{\linewidth}
        \centering
        \includegraphics[width=\linewidth]{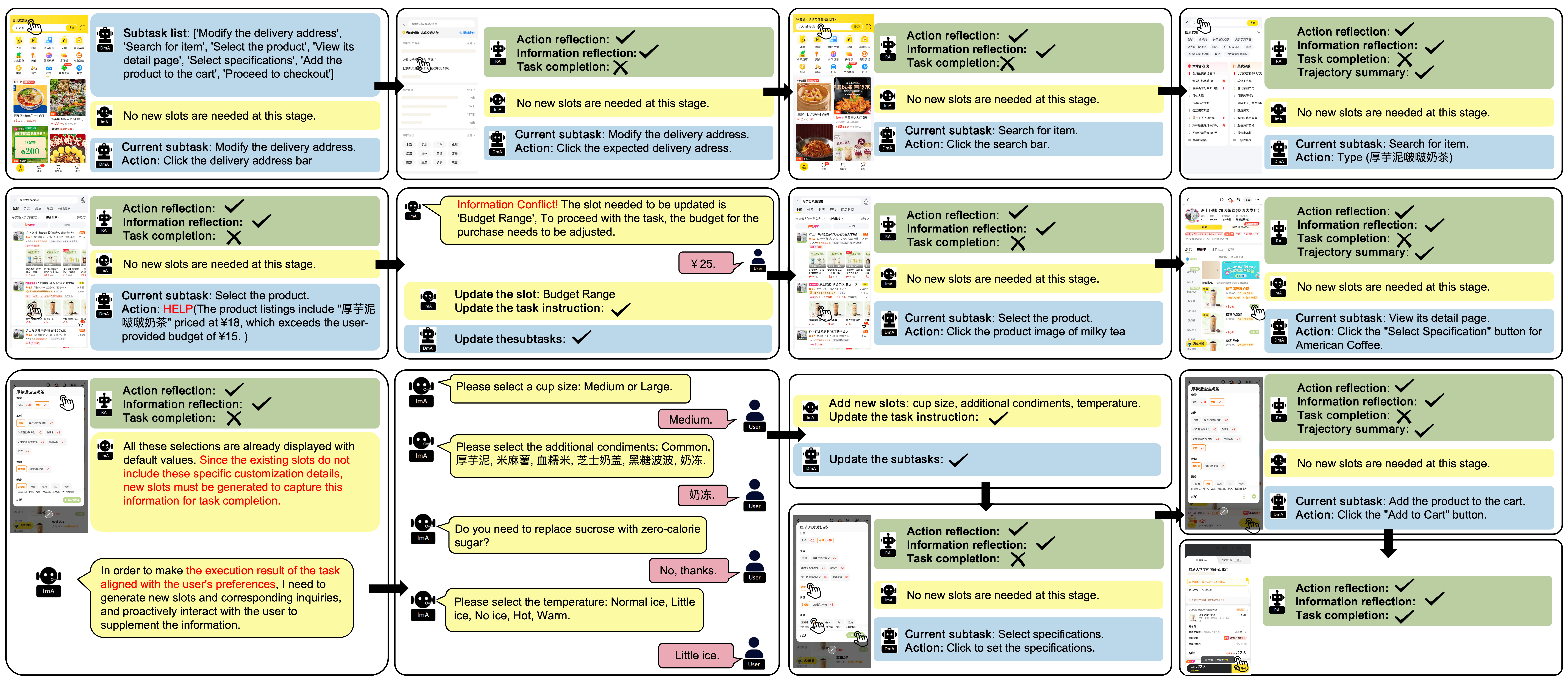}
        \caption{Stage 2: iterative task automation.}
        \label{casestudy:subfig2}
    \end{subfigure}
    
    \caption{A case of ReInAgent's two-stage processing in a food delivery task with three types of information dilemmas. The task is "Please help me order a cup of thick taro boba milk tea priced at no more than 15 RMB."}
    \label{casestudy}
\end{figure*}

\subsubsection{Evaluation on Different Tasks.} 
Table 2 presents ReInAgent’s performance across four task scenarios with app‑specific operational knowledge. SR and CR are notably higher in shopping online and takeaway ordering tasks. It achieves an SR of 14 out of 15 and a CR of 0.98 on JD.com, while reaching a SR of 14 out of 15 and a CR of 0.95 on Meituan, along with RE peaking at 0.83. Performance is less robust in the flight booking and hotel reservation tasks. The Chinese app Ctrip yields lower SR and CR than English apps like Skyscanner and Booking.com. Other metrics display no significant differences.

\begin{table}[t]
    \label{table3}
    \centering
    \small
    \renewcommand{\arraystretch}{1.2}
    \setlength{\tabcolsep}{3pt}
    \begin{tabular*}{\linewidth}{@{\extracolsep{\fill}}
        c
        >{\centering\arraybackslash}c
        >{\centering\arraybackslash}c
        cccc
        cc
        @{}}
    \toprule
    \multicolumn{2}{c}{\textbf{Ablation Module}}
      & \multicolumn{4}{c}{\textbf{Task-level}}
      & \multicolumn{2}{c}{\textbf{Info-level}} \\ 
    \cmidrule(l{0pt}r{2pt}){1-2} \cmidrule(l{2pt}r{2pt}){3-6} \cmidrule(l{2pt}r{4pt}){7-8}
    \makecell[c]{Action\\Reflection}
      & \makecell[c]{Information\\Reflection}
      & SR & CR & DA & RE & IC & CCR \\ 
    \midrule
      & \checkmark & 0.64 & 0.76 & 0.78 & 0.45 & 0.95 & \textbf{0.80} \\ 
    \checkmark &            & 0.78 & 0.87 & 0.83 & \textbf{0.79} & 0.77 & 0.56 \\ 
    \checkmark & \checkmark & \textbf{0.82} & \textbf{0.91} & \textbf{0.91} & 0.76 & \textbf{0.97} & 0.78 \\
    \bottomrule
    \end{tabular*}
    \caption{Ablation results of the action reflection and information reflection modules.}
\end{table}
\subsubsection{Ablation Study.}\label{ablation study}
We conduct ablation studies on the action reflection module and the information-consistency reflection module. Table 3 shows that omitting action reflection causes a pronounced drop in task-level performance. In the absence of action reflection, the decision-making agent repeats erroneous steps and fails to recover from incorrect interfaces, leading to flawed plans and inefficient progress. Removing information consistency reflection likewise diminishes output consistency and impairs the agent’s ability to detect and resolve conflicting information, causing it to overlook fine‐grained task details. These findings confirm that the two modules are essential for ReInAgent to deliver accurate and reliable task execution.

\subsubsection{Case Study.}
Figure~\ref{casestudy} illustrates the complete process of ReInAgent performing a food delivery task under the condition of injected app operation knowledge. In particular, Figure~\ref{casestudy:subfig1} demonstrates the task preprocessing stage, where the ImA engages the user to clarify ambiguous instructions. Figure~\ref{casestudy:subfig2} depicts the task execution stage, where the DmA generates a list of subtasks and then collaborates with the other agents in an iterative cycle to fulfill the user’s request.
During execution, the system encounters missing product specification details and a budget conflict. ReInAgent detects these dilemmas, prompts the user to specify preferences and adjust the budget, and then revises both the task instructions and the subtask list to enable dynamic evolution of the task. This case study highlights ReInAgent’s adaptability and flexibility in managing complex scenarios. Further cases for the remaining three task scenarios appear in Appendix.

\section{Conclusion and Future Work}\label{sec:conclusion and future work}
During mobile agent task automation, agents often face information dilemmas such as ambiguous instructions, incremental information supplementation, and conflicting requirements, which can severely impair execution and compromise outcome accuracy. To overcome these challenges and support dynamic task evolution, we introduce ReInAgent, a human-in-the-loop multi-agent framework with a dynamic task-slot management mechanism combined with proactive user-agent interaction. Experimental results show that ReInAgent effectively addresses information dilemmas and completes user tasks more accurately by enabling dynamic task evolution. However, ReInAgent may falter when handling screens with highly complex information, resulting in omission of critical incremental information and the failure to detect conflicts. In the future, we plan to frame information management as a partially observable Markov decision process and use reinforcement learning to train the ImA on optimal slot-querying and conflict-resolution policies. We will also perform extensive user-agent interaction simulations to systematically evaluate and refine the framework, enhancing its responsiveness to and management of complex information dilemmas.

\bibliography{aaai2026}

\clearpage
\appendix
\section{Appendix for Experiments}
\subsection{Effect of Trajectory Window Size}\label{appendix:Effect of trajectory window size}
\begin{figure}[H]
    \centering
    \includegraphics[width=\linewidth]{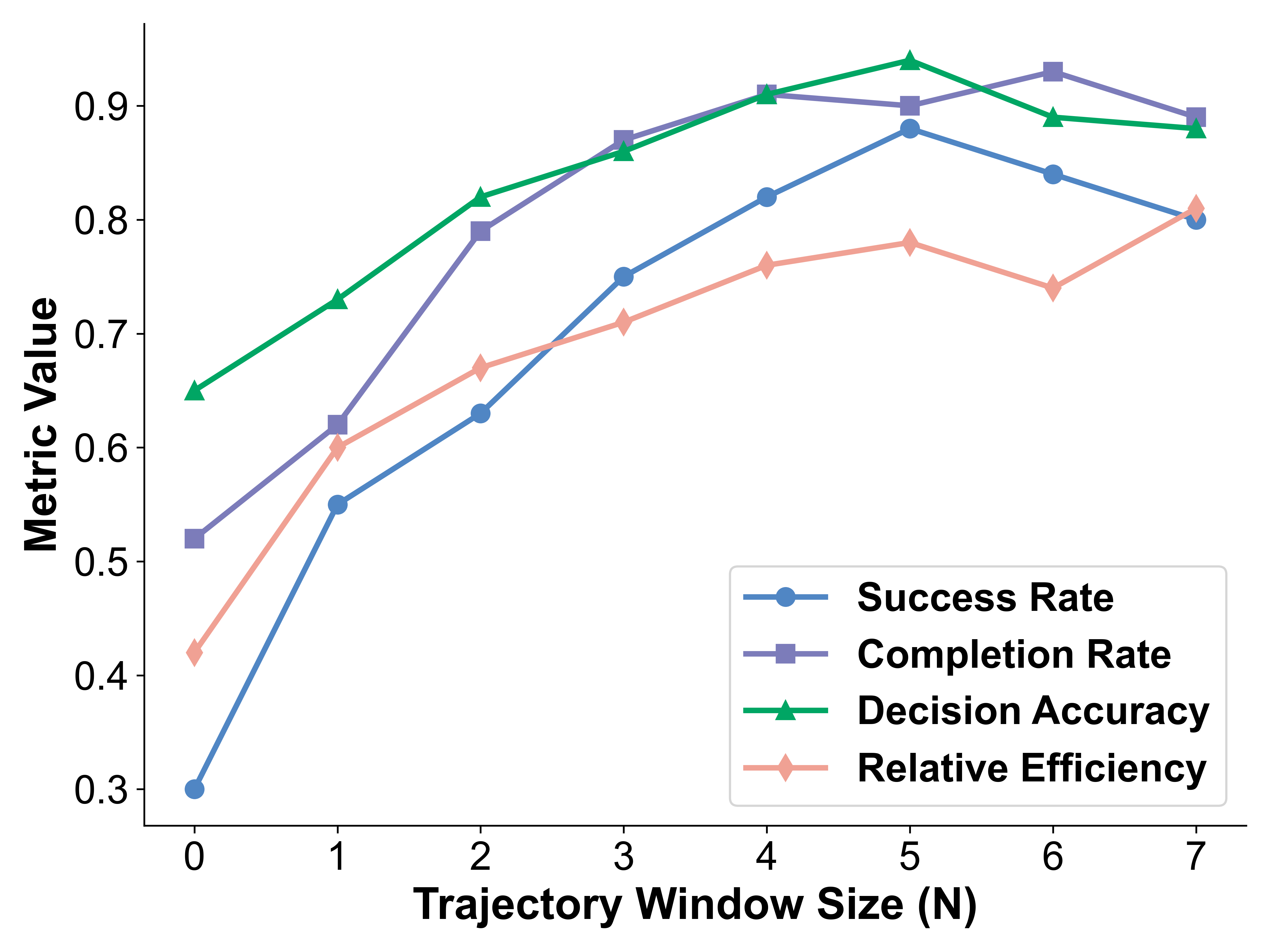}
    \caption{Influence of different trajectory window sizes.}
    \label{fig:trajectory window size}
\end{figure}

We evaluate the effect of the trajectory window size N on SR, CR, DA and RE, as shown in Figure~\ref{fig:trajectory window size}. When \(N \leq 1\), the agent lacks sufficient historical context, resulting in poor end-to-end performance. Performance improves steadily as \(N\) increases, with notable gains observed up to \(N=4\). Although DA and SR reach their peak at \(N=5\), we observe a slight decline and increased variability in all metrics when \(N \geq 5\). This suggests that excessively large windows introduce redundant information and computational overhead, leading to diminishing returns. This behavior indicates that excessively large trajectory windows introduce redundant information and computational overhead, producing diminishing returns. Therefore, we identify \(N=4\) as the optimal window size, balancing decision quality and computational efficiency in GUI-based task execution.

\subsection{Metrics}\label{appendix:metrics}
In order to conduct a more detailed evaluation of task performance, we design two metric levels: task-level and information-level. 

\subsubsection{Task-level Metrics.} Task-level metrics emphasize evaluating the task execution process, consisting of success rate (SR), completion rate (CR), decision accuracy (DA), reflection accuracy (RA), action effectiveness (AE), and relative efficiency (RE). While SR refers to the proportion of fully completed tasks, CR indicates the proportion of accomplished subtasks. DA assesses correctness in understanding screen states, planning, and action decisions. RA evaluates correctness in action reflection, information consistency reflection, and overall task progress reflection. AE denotes the proportion of successfully executed actions on mobile devices. RE measures the ratio of the number of steps in an ideal human-executed scenario to those actually executed by the agent. 

\subsubsection{Information-level Metrics.} Information-level metrics focus on evaluating task-related information, including information consistency (IC), information capture rate (ICR), slot accuracy (SA), and conflict capture rate (CCR). IC assesses the alignment of the task execution result with user preferences. ICR is the percentage of task slots the agent correctly identifies. CCR denotes the fraction of information conflicts it successfully detects. SA represents the accuracy of slots generated during execution.

\subsection{Model Settings}\label{appendix:Model Settings}
We adopt GPT-4o as the foundational model for all three agents, utilizing it through API-based inference calls. When using GPT‑4o, we set its temperature at 0.2 as this low‑but‑non‑zero setting keeps the agent’s outputs almost deterministic, which minimizes hallucinations, mis‑clicks, and run‑to‑run drift, while preserving randomness to achieve two benefits: (i) enable the DmA to explore alternative plans and recover from dead ends according to reflection, and (ii) help the ImA generate flexible slots and natural follow‑up queries rather than rigid, repetitive forms. This balance between reliability and limited randomness is essential for ReInAgent’s robust performance across diverse mobile‑task scenarios. We also set the max-tokens at 2048, and other hyperparameters are set to default.
When using the OS-Atlas-7B model as a localization tool, we adopt its default parameter configuration and only modify the output format through prompt engineering.

\subsection{Additional Case Studies}\label{more cases}
\begin{figure*}[htbp]
    \centering
    \begin{subfigure}[b]{0.5\textwidth}
        \centering
        \includegraphics[width=\textwidth]{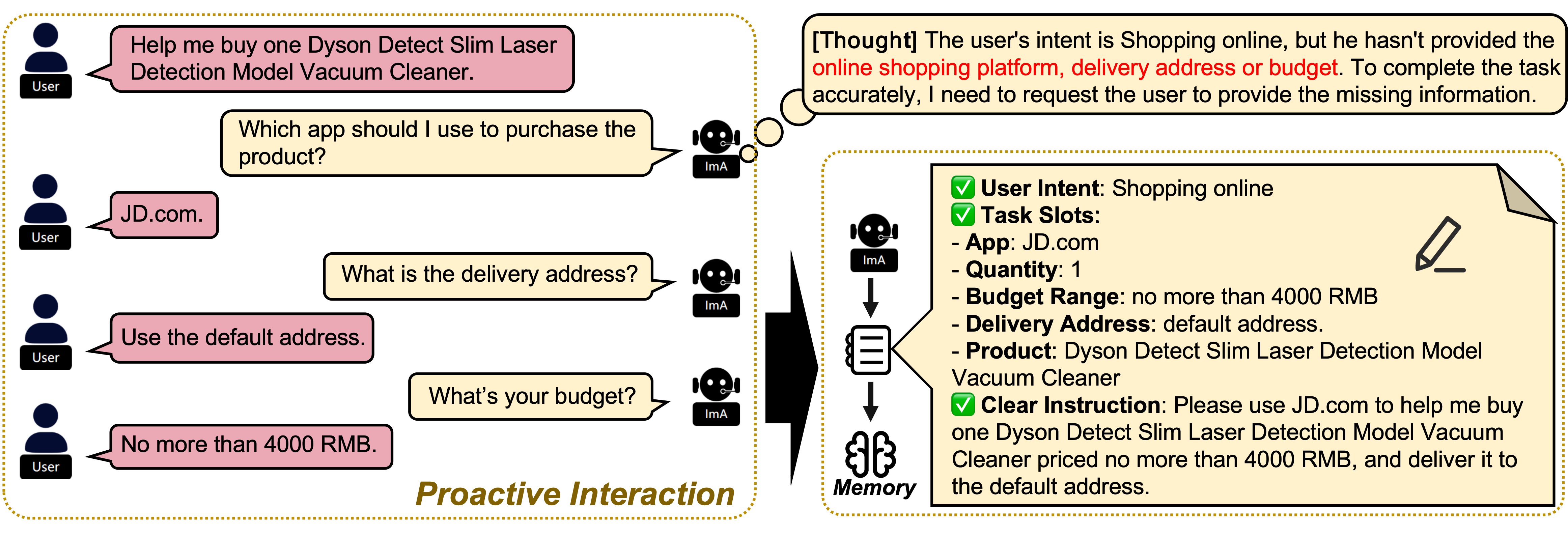}
        \caption{Stage 1: task pre-processing.}
        \label{casestudy:subfig1}
    \end{subfigure}

    \begin{subfigure}[b]{0.7\textwidth}
        \centering
        \includegraphics[width=\textwidth]{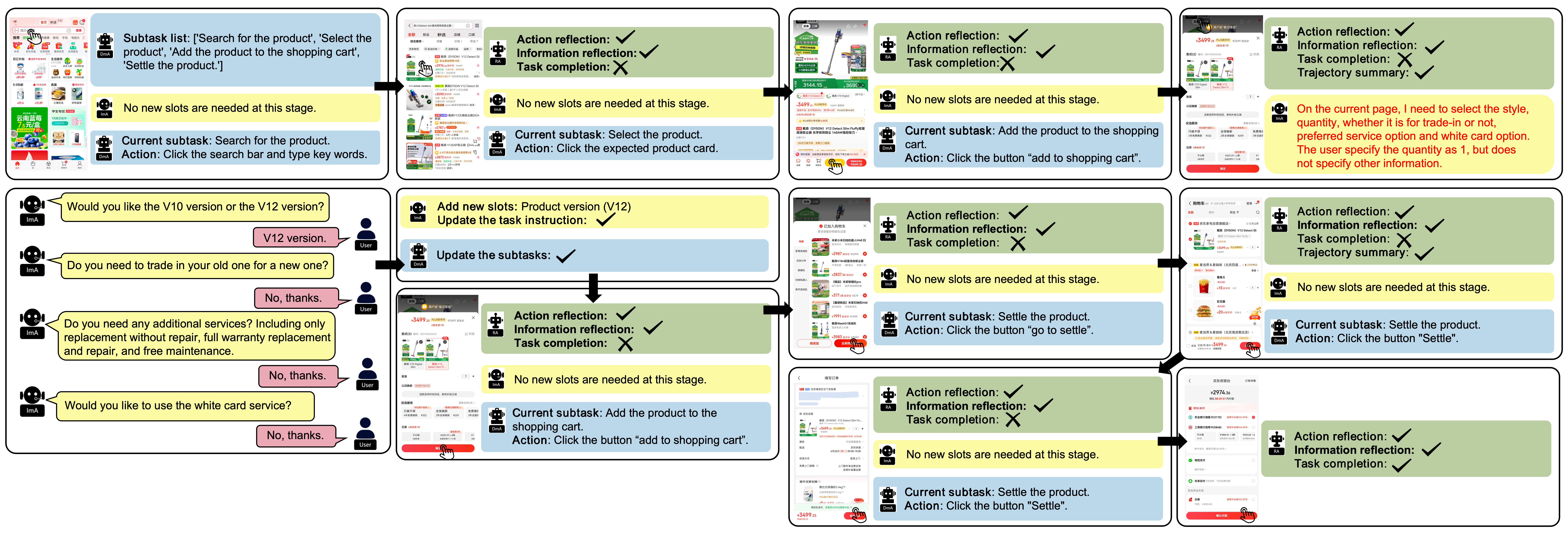}
        \caption{Stage 2: iterative task automation.}
        \label{casestudy:subfig2}
    \end{subfigure}
    
    \caption{A case of shopping online. The task is: "Help me buy one Dyson Detect Slim Laser Detection Model Vacuum Cleaner."}
    \label{fig:case_shopping}
\end{figure*}

\begin{figure*}[htbp]
    \centering
    \begin{subfigure}[b]{0.5\textwidth}
        \centering
        \includegraphics[width=\textwidth]{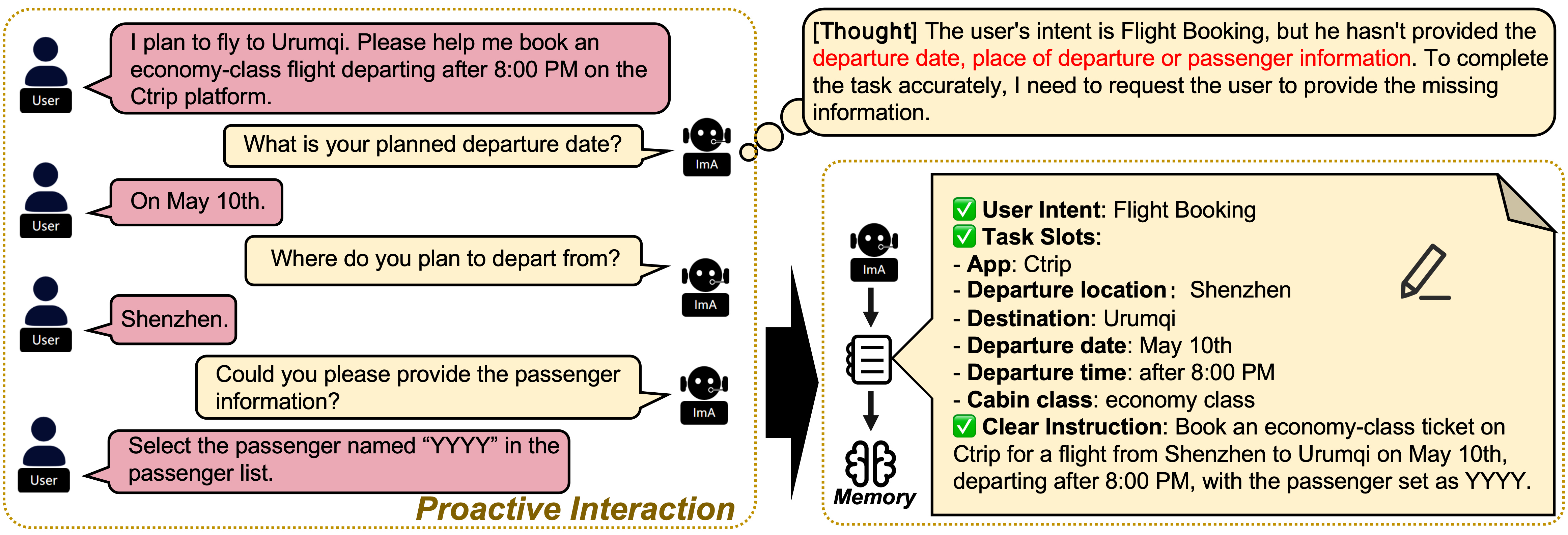}
        \caption{Stage 1: task pre-processing.}
        \label{casestudy:subfig1}
    \end{subfigure}

    \vspace{0.1cm}  

    \begin{subfigure}[b]{0.7\textwidth}
        \centering
        \includegraphics[width=\textwidth]{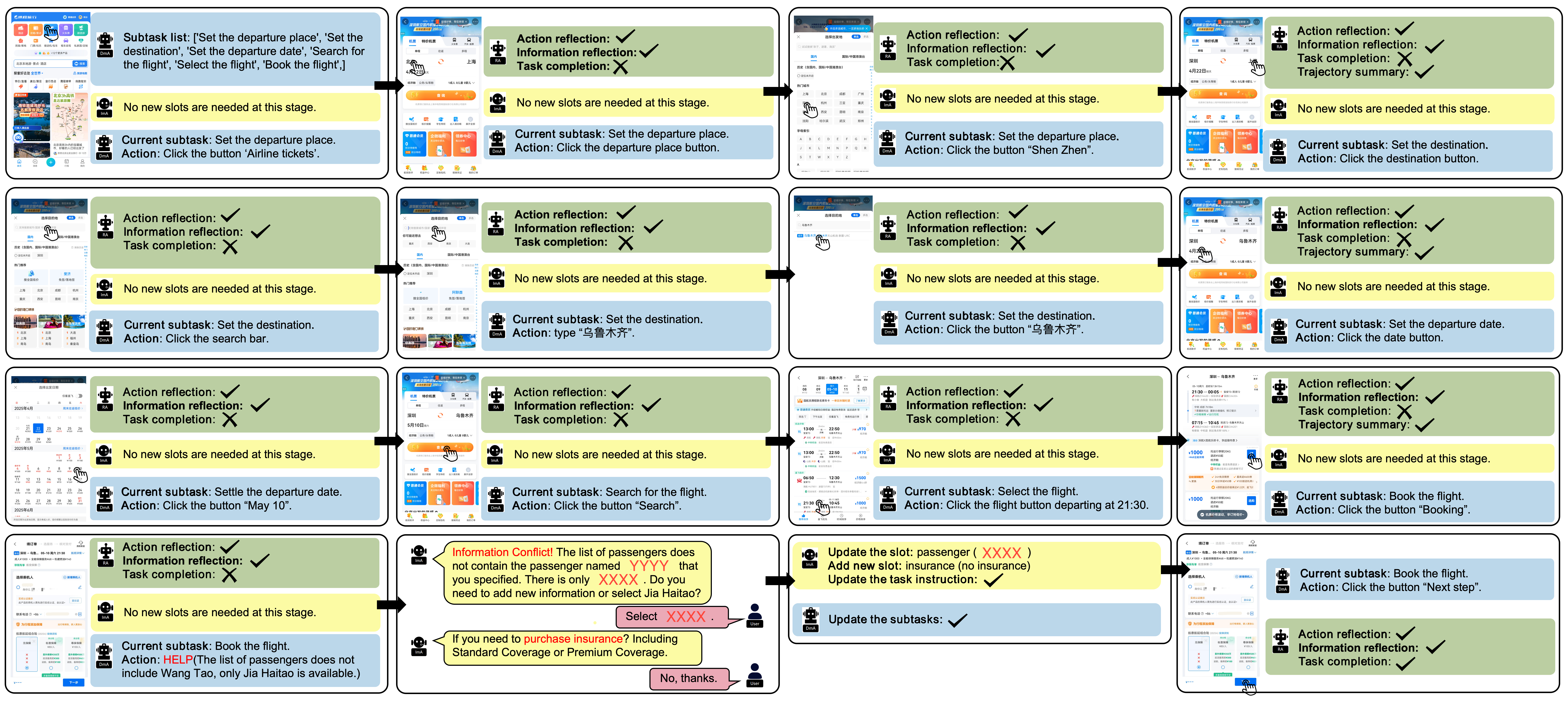}
        \caption{Stage 2: iterative task automation.}
        \label{casestudy:subfig2}
    \end{subfigure}
    
    \caption{A case of flight booking. The task is: "I plan to fly to Urumqi, please help me book an economy-class flight departing after 8:00 PM on the Ctrip platform."}
    \label{fig:case_flight}
\end{figure*}

\begin{figure*}[htbp]
    \vspace*{\fill}
    \centering
    \begin{subfigure}[b]{0.5\textwidth}
        \centering
        \includegraphics[width=\textwidth]{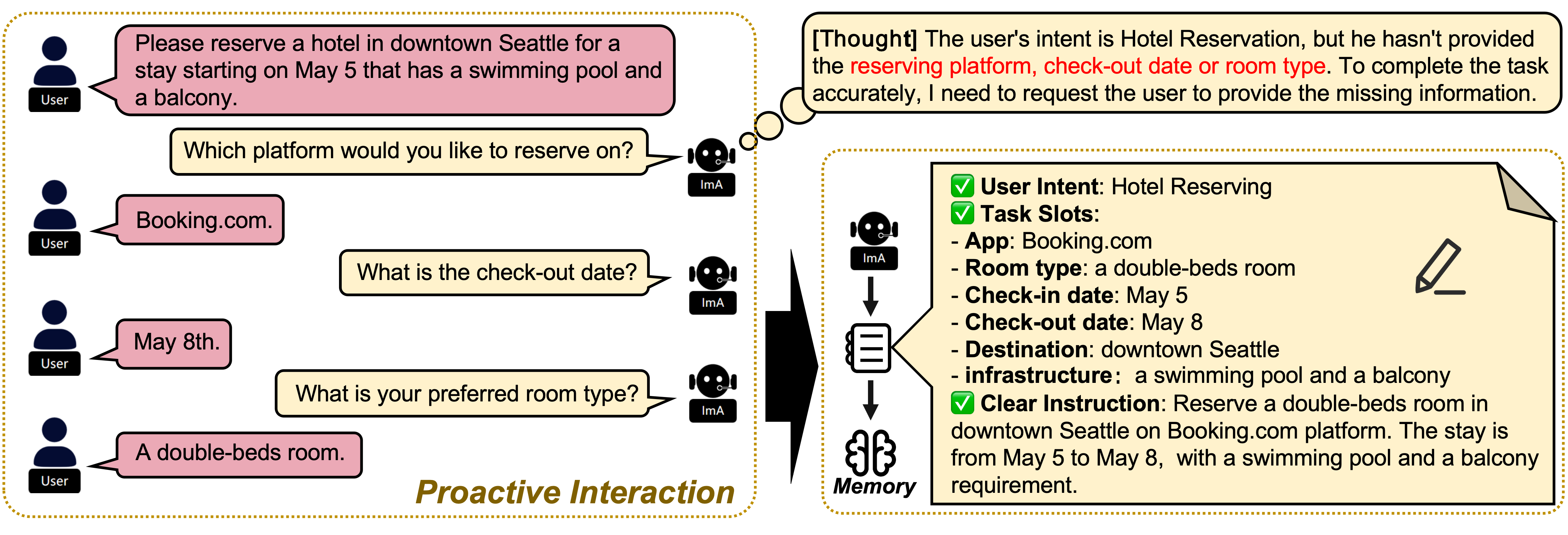}
        \caption{Stage 1: task pre-processing.}
        \label{casestudy:subfig1}
    \end{subfigure}

    \vspace{0.1cm}  

    \begin{subfigure}[b]{0.7\textwidth}
        \centering
        \includegraphics[width=\textwidth]{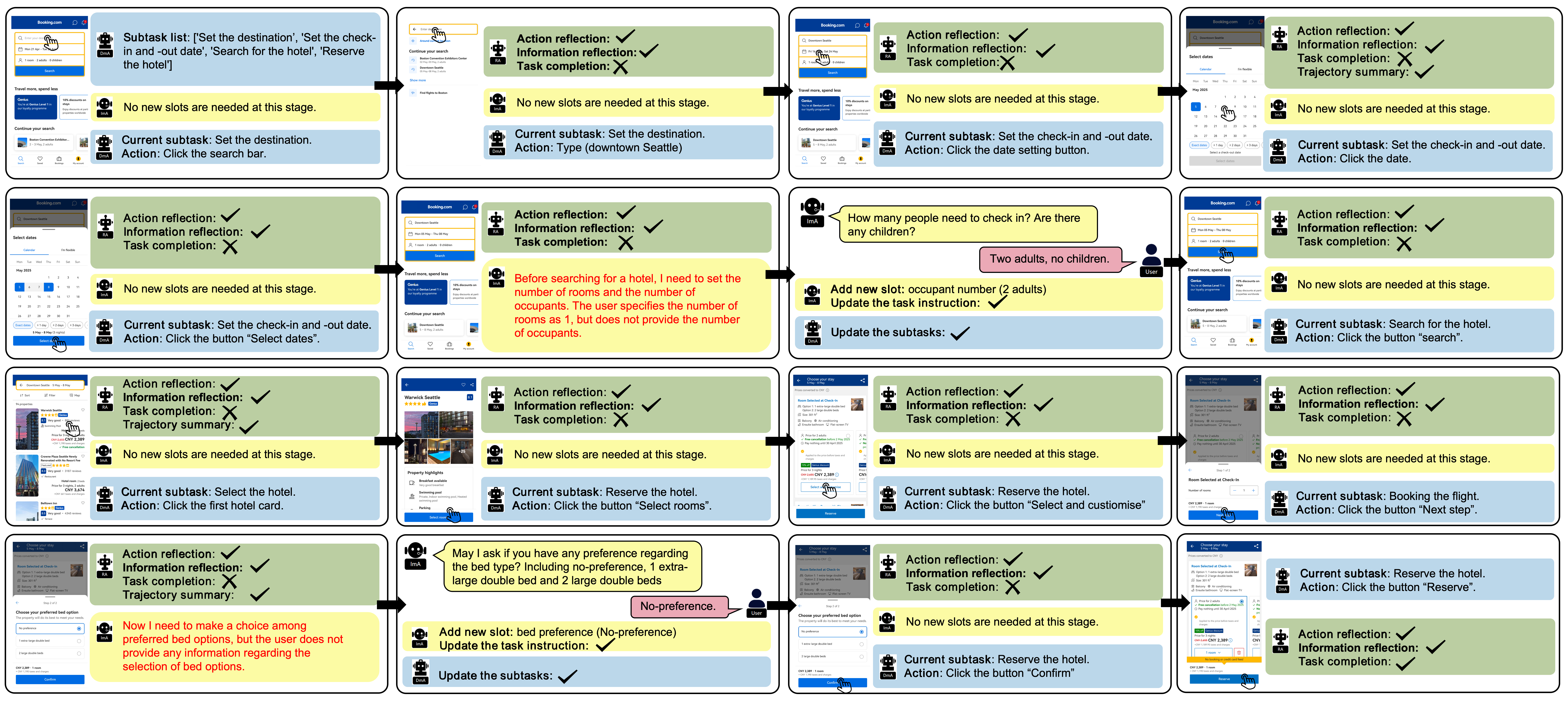}
        \caption{Stage 2: iterative task automation.}
        \label{casestudy:subfig2}
    \end{subfigure}
    
    \caption{A case of hotel reservation. The task is: "Reserve the hotel in downtown Seattle. The stay starts on May 5, with a swimming pool and a balcony requirement."}
    \label{fig:case_hotel}
\end{figure*}

As shown in Figures~\ref{fig:case_shopping},~\ref{fig:case_flight}, and~\ref{fig:case_hotel}, which respectively depict cases of information dilemmas in online shopping, flight booking, and hotel reservation scenarios, these examples demonstrate that ReInAgent can proactively interact with the user to effectively address the three types of information dilemmas throughout the task lifecycle, and that whenever slots are added or modified, the task context is updated simultaneously to achieve dynamic task evolution.

\subsection{Evaluation Tasks}\label{evaluation tasks}
We design evaluations for four types of complex tasks commonly found on mobile platforms: flight booking, takeaway ordering, hotel reservation, and online shopping. To assess our framework’s task execution capabilities across apps in different languages, we conduct comparative evaluations using both Chinese and English apps for the flight booking and hotel reservation tasks. Additionally, to highlight the three types of information dilemmas that agents may encounter during automated task execution, which are the focus of this paper, we deliberately design tasks that incorporate these dilemmas. The task instructions are provided in the code file.

\section{More Details on ReInAgent}

\subsection{Adaptive Action Space}\label{action space}
To simulate natural human–phone interaction, we define an action space \(\mathcal{A}\) for the DmA, encompassing fundamental mobile operations as illustrated in the Table 1. Notably, we introduce a meta-action, HELP, which enables the DmA to request assistance from the ImA. Whenever information dilemmas arise, the DmA autonomously generates a description of the dilemma and executes the HELP action to solicit guidance from the ImA.

\begin{table}[htbp]

    \label{tab: Adaptive action space}
    \centering
    \small
    \renewcommand{\arraystretch}{1.2}
    \begin{tabularx}{\linewidth}{lX}
        \toprule
        \textbf{Action} & \textbf{Description} \\
        \midrule
        CLICK (content) & Performs a click operation within the current screen. The \textit{content} describes the control element to be interacted with. \\
        SLIDE (direction) & Performs a sliding operation on the screen. The \textit{direction} specifies the sliding direction. \\
        TYPE (text) & Inputs textual information. The \textit{text} specifies the exact content to be entered. \\
        BACK & Returns to the previous screen. \\
        HOME & Returns to the mobile device's home screen. \\
        HELP (dilemma) & Requests assistance from the ImA. The \textit{dilemma} describes the current informational challenge, including conflicts of information or sensitive operations. \\
        \bottomrule
    \end{tabularx}
    \caption{Adaptive action space.}
\end{table}

\subsection{Prompts for the ReInAgent}\label{appendix:prompt}
When we design ReInAgent’s prompts, we leverage in-context learning and chain-of-thought techniques to steer the LLM’s reasoning and output generation, yielding more precise and contextually appropriate responses. At the same time, we specify an XML-based output schema within the prompts, enforcing a consistent, machine-readable format that facilitates downstream data processing. The prompt configurations for the information-managing agent’s appear in Tables 2, 3, 4, 5, 6, and 7; those for the reflecting agent are given in Tables 8 and 9; and the decision-making agent templates are listed in Tables 10, 11, and 12.

\begin{table*}[htbp]

\label{tab:Prompt for ImA's intent recognition.}
\centering
\renewcommand{\arraystretch}{0.9}  
\begin{tabular}{>{\arraybackslash\hspace{0.5em}}p{\dimexpr\textwidth-2\tabcolsep-2\arrayrulewidth}<{\arraybackslash\hspace{0.5em}\scriptsize}}
\toprule
\textbf{ImA: Intent Recognition}\\
\midrule
\#\# Role and Task\\
    I am an intent recognition agent within a mobile intelligent framework. My task is to receive and understand the user's initial task instruction and determine the intent by selecting the most appropriate intent type from the predefined list of user intents.\\\\
    \#\# User Intents\\
    - Flight Booking\\
    - Takeaway Delivery\\
    - Hotel Reservation\\
    - Online Shopping\\\\
    \#\# Output Format\\
    \textless  intent\textgreater [Fill in intent]\textless  /intent\textgreater \\\\
    \#\# Input\\
    - User's initial task instruction: \{taskIns\}\\
\bottomrule
\end{tabular}
\caption{Prompt for ImA's intent recognition.}
\end{table*}

\begin{table*}[htbp]

\label{tab:Prompt for ImA's instruction updating.}
\centering
\renewcommand{\arraystretch}{0.9}  
\begin{tabular}{>{\arraybackslash\hspace{0.5em}}p{\dimexpr\textwidth-2\tabcolsep-2\arrayrulewidth}<{\arraybackslash\hspace{0.5em}\scriptsize}}
\toprule
\textbf{ImA: Instruction Updating}\\
\midrule
\#\# Role and Task\\
I am an information processing agent within a mobile intelligent framework. My current task is to update the user's initial task instruction using the supplemented complete slot information, generating a task instruction that includes all necessary slot details.\\\\
    
    \#\# Output Format\\
    \textless  updated\_task\_ins \textgreater [Fill in updated\_task\_ins]\textless  /updated\_task\_ins \textgreater \\\\
    
    \#\# Input\\
    - User's initial task instruction: \{taskIns\}\\
    - Supplemented complete slot information: [\{slots\_info\}]\\
\bottomrule
\end{tabular}
\caption{Prompt for ImA's instruction updating.}
\end{table*}

\begin{table*}[htbp]

\label{tab:The first part of the prompt for ImA's dynamical slots generation.}
\centering
\renewcommand{\arraystretch}{0.9}  
\begin{tabular}{>{\arraybackslash\hspace{0.5em}}p{\dimexpr\textwidth-2\tabcolsep-2\arrayrulewidth}<{\arraybackslash\hspace{0.5em}\scriptsize}}
\toprule
\textbf{ImA: Dynamical Slots Generation Part 1}\\
\midrule
\#\# Role Positioning\\
    You are a slot maintenance expert in a mobile intelligent assistant framework. Before the downstream assistant operates the mobile device, your task is to analyze the mobile screenshot, understand the user’s task, and determine whether new slots need to be generated. These slots will be referenced by other collaborating agents.\\\\
    
    \#\# Inference Steps\\
    1. Analyze the mobile screenshot in the context of the user’s task and interpret the current page state.\\
    2. Expand the semantic meaning of existing slots, for example:\\
       - Food Selection → can map to Product Name/Search Keyword\\
       - Contact Information → can map to Phone Number/Mobile Number\\
       - Address Information → can map to Delivery Address/Pickup Location\\
    3. Establish cross-field logical mappings. For example, if a search keyword is needed and existing slots contain specific product details (e.g., food selection, product type, brand preference), a new slot should not be generated.\\
    4. Validate the following conditions step by step to determine whether a new slot is needed:\\
       - 4.1 If the page requires input but the required information cannot be mapped to an existing slot name, generate a new slot.\\
         - Example: If an order needs to be placed and the page requires a phone number, but the existing slots do not include a phone number, generate a new slot 'Phone Number'.\\
       - 4.2 If the page requires input and already has a default value, but the existing slots do not include that information, generate a new slot and ask the user for confirmation.\\
         - Example: If the default date on the page is 'February 1', but there is no 'Date' slot, generate a new slot 'Date'.\\
       - 4.3 If the page requires selecting with default options already chosen, but no corresponding slot guides this selection, a slot must be generated.\\
         - Example: If the page allows selecting cup size (Large, Medium, Small) and temperature settings, and the default selection is 'Small Cup, Less Ice', but the existing slots only contain 'temperature settings: Less Ice', a new slot 'Cup Size' must be generated.\\\\
\bottomrule
\end{tabular}
\caption{The first part of the prompt for ImA's dynamical slots generation.}
\end{table*}

\begin{table*}[htbp]

\label{tab:The second part of the prompt for ImA's dynamical slots generation.}
\centering
\renewcommand{\arraystretch}{1}  
\begin{tabular}{> {\arraybackslash\hspace{0.5em}}p{\dimexpr\textwidth-2\tabcolsep-2\arrayrulewidth}<{\arraybackslash\hspace{0.5em}\scriptsize}}
\toprule
\textbf{ImA: Dynamical Slots Generation Part 2}\\
\midrule
    \#\# Slot Generation Requirements\\
    - Slot Naming: Must be 2-5 words, using clear nouns. Sentences or vague descriptions are prohibited.\\
    - Information Association: Must directly correspond to the currently focused input area or selection control but should not involve non-input actions like clicking search.\\
    - Selection Guidance: If multiple options exist on the page (e.g., flight time slots), and the slot does not exist, generate a new slot 'Flight Time' to guide the selection process.\\
    - Ignore default selections on the page and generate slots for each category requiring selection.\\\\
    \#\# Slot Generation Restrictions\\
        - Do not consider whether the current page state matches existing slot information.\\
        - Do not generate a slot similar to the existing slot information.\\
        - If a slot’s value can be inferred from existing slots, do not generate a new slot. \\
          - Example: If the 'Food Selection' slot contains 'Coffee', and the page requires entering a search keyword, use the existing slot value instead of generating a new 'Search Keyword' slot.\\
        - If a slot can be mapped to an existing slot, use the existing one instead of creating a new one.\\
        - Do not create new slots based only on missing textual information—semantic equivalence must be considered.\\
        - Do not create new slots to override existing slot values. Example: If a 'Delivery Address' slot already exists but is not displayed on the page, do not generate a new 'Address' slot.\\
        - Do not generate slots to confirm operations.\\
        - If there is no selection or input requirement for the current page, there is no need to predict the future state that may need to generate slots, only the necessary slots are generated for the current page state.
    \\
    \#\# Output Format\\
    The response format must be strictly limited to one of the following:\\
    1. If new slots need to be generated, output the following format for each slot:\\
    ```xml\\
    \textless  screen\_description\textgreater [Description of screenshot content related to the user’s task]\textless  /screen\_description\textgreater 
    \textless  thought\textgreater [Reasoning on whether a new slot is needed, reflection on slot generation restrictions, and final decision on generating a slot]\textless  /thought \textgreater 
    \textless  num \textgreater [Number of new slots to be generated]\textless  /num \textgreater 
    \textless slot\_name\textgreater [New slot name]\textless /slot\_name\textgreater 
    \textless inquiry\textgreater [Request for missing slot information; if multiple options exist, list them in the inquiry]\textless /inquiry\textgreater \\
    ```\\
    Example Output:\\
    ```xml\\
    \textless screen\_description\textgreater This screenshot displays an order details page. The user task is food delivery, requiring order information completion before placing the order. The page shows the username as 'Zhou Wang', but the phone number and address are missing. A 'Checkout' button is available at the bottom.\textless /screen\_description\textgreater 
    \textless thought\textgreater The page requires a phone number and address, but the existing slots do not contain this information. Therefore, these two new slots must be generated to complete the subtask.\textless /thought\textgreater 
    \textless num\textgreater 2\textless /num\textgreater 
    \textless slot\_name\textgreater Phone Number\textless /slot\_name\textgreater 
    \textless inquiry\textgreater The current page requires entering a phone number. Please provide your phone number.\textless /inquiry\textgreater 
    \textless slot\_name\textgreater Address\textless /slot\_name\textgreater 
    \textless inquiry\textgreater The current page requires selecting a delivery address. Please provide your preferred address.\textless /inquiry\textgreater \\
    ```\\
    2. If no new slots are needed, output the following format:\\
    ```xml\\
    \textless screen\_description\textgreater [Description of screenshot content related to the user’s task]\textless /screen\_description\textgreater 
    \textless thought\textgreater [Reasoning on whether a new slot is needed, reflection on slot generation restrictions, and final decision on generating a slot]\textless /thought\textgreater 
    \textless num\textgreater 0\textless /num\textgreater \\
    ```\\\\
    \#\# Input\\
    - User Task: \{task\}\\
    - Existing Slots: [\{slots\_info\}]\\
    - Mobile Screenshot: (as shown in the image)\\
    Before generating output, review the slot generation requirements and restrictions again. Your focus is solely on whether the page requires information input and whether the required information exists in the current slots. You do not need to consider how to execute the task or how to interact with the interface.

\\
\bottomrule
\end{tabular}
\caption{The second part of the prompt for ImA's dynamical slots generation.}
\end{table*}

\begin{table*}[htbp]

\label{tab:Prompt for ImA's instruction preprocessing.}
\centering
\renewcommand{\arraystretch}{0.9}  
\begin{tabular}{>{\arraybackslash\hspace{0.5em}}p{\dimexpr\textwidth-2\tabcolsep-2\arrayrulewidth}<{\arraybackslash\hspace{0.5em}\scriptsize}}
\toprule
\textbf{ImA: Task Instruction Pre-processing}\\
\midrule
\#\# Role Positioning\\
    I am an instruction clarification agent in a mobile intelligent framework. My tasks are:\\
    1. Receive and understand the user's initial task instruction to identify the intent.\\
    2. Extract all explicitly mentioned task details from the user instruction and generate corresponding slots.\\
    3. Infer and generate all necessary slots required to complete the task, even if not mentioned in the user instruction, ensuring all essential information is available for task execution.\\
    4. If some slot values cannot be extracted from the user instruction, generate targeted inquiries requesting user input.\\
    
    \#\# Step-by-step Processing:\\
    - Step 1: Parse the user instruction and extract all explicitly mentioned task details as slots.\\
    - Step 2: Based on the task intent, supplement the necessary slots required to complete the task (e.g., time, location, quantity, preferences) even if they are not mentioned.\\
    - Step 3: For all slots where the value cannot be extracted, generate a targeted inquiry.\\
    
    \#\# Output Format
    For each slot, output:\\
    ```xml\\
    \textless num\textgreater [Total number of slots]\textless /num\textgreater 
    \textless slot\_name\textgreater [Slot Name]\textless /slot\_name\textgreater \textless slot\_value\textgreater [Extracted Slot Value, or null if not found]\textless /slot\_value\textgreater \textless inquiry\textgreater [If slot\_value is null, generate a targeted inquiry; otherwise, null]\textless /inquiry\textgreater 
    ...
    \textless slot\_name\textgreater app\textless /slot\_name\textgreater \textless slot\_value\textgreater [App Name]\textless /slot\_value\textgreater \textless inquiry\textgreater null\textless /inquiry\textgreater \\
    ```\\
    
    \#\# Requirements\\
    - The first slot must be `app`, indicating the platform for task execution. If the user instruction does not specify the app, an inquiry must be generated.\\
    - Every explicitly mentioned task detail in the user instruction must have a corresponding slot.\\
    - Generate all necessary slots based on task intent, even if not mentioned by the user. Essential slots include but are not limited to:\\
      - For booking/transportation tasks: Departure location, destination, departure time, vehicle/service type, budget range, passenger count, etc.\\
      - For search tasks: Keywords, time range, search platform, etc.\\
      - For purchase tasks: Product name, quantity, payment method, budget range, etc.\\
      - For other tasks: Context-based necessary parameters such as time, location, quantity, user preferences, etc.\\
    - If any essential slot for fulfilling the task intent is missing, it must be generated with a null value and a corresponding inquiry.\\
    - Generated inquiries must be clear and specific to guide the user in providing the missing information.\\
    
    \#\# Example\\
    Input:\\
    taskIns: Reserve a Didi from Tianjin Binhai Airport to 1 Shuishang Park East Road, Nankai.\\
    intent: Online Ride Booking\\
    
    Output:\\
    ```xml\\
    \textless num\textgreater 6\textless /num\textgreater \\
    \textless slot\_name\textgreater app\textless /slot\_name\textgreater \textless slot\_value\textgreater Didi\textless /slot\_value\textgreater \textless inquiry\textgreater null\textless /inquiry\textgreater \\
    \textless slot\_name\textgreater Departure Location\textless /slot\_name\textgreater \textless slot\_value\textgreater Tianjin Binhai Airport\textless /slot\_value\textgreater \textless inquiry\textgreater null\textless /inquiry\textgreater \\
    \textless slot\_name\textgreater Destination\textless /slot\_name\textgreater \textless slot\_value\textgreater 1 Shuishang Park East Road, Nankai\textless /slot\_value\textgreater \textless inquiry\textgreater null\textless /inquiry\textgreater \\
    \textless slot\_name\textgreater Departure Time\textless /slot\_name\textgreater \textless slot\_value\textgreater null\textless /slot\_value\textgreater \textless inquiry\textgreater What time do you need to book the ride?\textless /inquiry\textgreater \\
    \textless slot\_name\textgreater Vehicle Type\textless /slot\_name\textgreater \textless slot\_value\textgreater null\textless /slot\_value\textgreater \textless inquiry\textgreater Which vehicle type do you prefer (e.g., economy, premium)?\textless /inquiry\textgreater \\
    \textless slot\_name\textgreater Budget Range\textless /slot\_name\textgreater \textless slot\_value\textgreater null\textless /slot\_value\textgreater \textless inquiry\textgreater What is your budget range?\textless /inquiry\textgreater \\
    ```\\
    
    \#\# Input\\
    The user's initial task instruction is [\{taskIns\}], and the task intent is [\{intent\}].
\\
\bottomrule
\end{tabular}
\caption{Prompt for ImA's instruction pre-processing.}
\end{table*}

\begin{table*}[htbp]

\label{tab:Prompt for ImA's dilemma classification.}
\centering
\renewcommand{\arraystretch}{0.9}  
\begin{tabular}{>{\arraybackslash\hspace{0.5em}}p{\dimexpr\textwidth-2\tabcolsep-2\arrayrulewidth}<{\arraybackslash\hspace{0.5em}\scriptsize}}
\toprule
\textbf{ImA: Dilemma Classification}\\
\midrule
    \#\# Role Definition\\
    - Role: Information Dilemma Handling Expert\\
    - Task: Receive dilemma descriptions from the decision agent, accurately classify the dilemma based on a predefined categorization process using the user task instructions, current screen state, and existing slot information, and resolve the dilemma accordingly using the corresponding handling strategy. The response must follow the specified output format.\\\\
    
    \#\# Predefined Dilemma Categories\\
    - Dilemma 1: Slot Information Conflict – The current page’s information state conflicts with the task instructions and existing slot requirements, preventing further execution as per user instructions. This dilemma can be resolved by modifying slot information. Examples include:\\
      - Value Conflicts (e.g., price/date exceeding limits)\\
      - State Conflicts (e.g., out-of-stock products, sold-out seats, unsearchable/nonexistent items)\\
      - Logical Conflicts (e.g., number of child tickets exceeding adult tickets)\\
    
    - Dilemma 2: User Privacy \& Security – Involves sensitive operations (login, verification, payment) and private information (name, phone number, ID number, password), or repeated failed attempts with the same action that still fail to reach the goal. This dilemma requires the user to directly operate the screen.\\\\
    
    \#\# Predefined Classification Process\\
    1. Does the dilemma description mention login, verification, payment, repeated failed operations, or request user intervention?\\
       - Yes to Classify as Dilemma 2.\\
       - Example: "The current page requires checkout" to Classified as Dilemma 2.\\
       
    2. Can the dilemma be resolved by modifying existing slot information?\\
       - Yes to Classify as Dilemma 1.\\
       - Example: "Multiple search attempts failed to find the specified product, but modifying the 'product name' slot could resolve it" to Classified as Dilemma 1.\\\\
    
    \#\# Dilemma Handling Rules\\
    \#\#\# Dilemma 1 (Slot Information Conflict):\\
    - Handling Strategy: Identify the specific slot that needs to be updated based on user task instructions, the current screen state, dilemma description, and existing slot information.\\
    - Output Content:\\
      thought: Reasoning and detailed understanding of the dilemma, including constraints.;
      update\_info\_name: The name of the slot that needs to be updated;
      inquiry: A direct request to update the slot, specifying the necessary changes and reasoning without uncertainty (e.g., avoid phrases like "Do you confirm?").\\
    - Output Format:\\
      '''xml\\
      \textless dilemma\_type\textgreater Dilemma 1\textless /dilemma\_type\textgreater 
      \textless thought\textgreater [Fill in thought]\textless /thought\textgreater 
      \textless update\_info\_name\textgreater [Fill in update\_info\_name]\textless /update\_info\_name\textgreater 
      \textless inquiry\textgreater [Fill in inquiry]\textless /inquiry\textgreater 
      '''\\
    
    \#\#\# Dilemma 2 (User Privacy \& Security):\\
    - Handling Strategy: Request the user to directly operate the mobile screen.\\
    - Output Content:\\
      reason: Explanation of the dilemma;
      inquiry: Request for the user to take direct action on the mobile screen.\\
    - Output Format:\\
      '''xml\\
      \textless dilemma\_type\textgreater Operational Dilemma\textless /dilemma\_type\textgreater 
      \textless reason\textgreater [Fill in reason]\textless /reason\textgreater 
      \textless inquiry\textgreater [Fill in inquiry]\textless /inquiry\textgreater \\
      '''\\\\
    
    \#\# Input\\
    - Dilemma description: [\{dilemma\_description\}]\\
    - User task instruction: [\{task\_ins\}]\\
    - Current slot information: [\{slot\_info\}]\\
    - Current screen state: (as shown in the image)\\
\\
\bottomrule
\end{tabular}
\caption{Prompt for ImA's dilemma classification.}
\end{table*}

\begin{table*}[htbp]

\label{tab:Prompt for RA's reflection.}
\centering
\renewcommand{\arraystretch}{1.4}  
\begin{tabular}{>{\arraybackslash\hspace{0.5em}}p{\dimexpr\textwidth-2\tabcolsep-2\arrayrulewidth}<{\arraybackslash\hspace{0.5em}\scriptsize}}
\toprule
\textbf{RA: Reflecting}\\
\midrule
\#\# Role and Task\\
    I am a reflection expert for mobile operations. My tasks are:\\
    1. First, analyze the previous action, the screenshot before execution, and the screenshot after execution to determine whether the action achieved the expected goal.\\
    2. If the previous action met its goal, use the post-execution screenshot and task progress to assess whether the complete user instruction task is finished (typically, when all subtasks in the subtask list are completed, the user instruction task is considered finished).\\\\
    
    \#\# Action Reflection Rules\\
    - When the action is clicking an input field, assume the action content is valid.\\
    - When the action is 'HELP', assume the action content is valid.\\
    - If the executed action results in navigating to an unexpected page, consider the action invalid and require exiting the page to retry.\\\\
    
    \#\# output\\
    - actionReflection: Includes pre-action screen state, action, post-action screen state, whether the expected page was reached, whether the action met its planned goal, and whether the action aligns with user task instructions, slot requirements, and app operation knowledge.\\
    - whether\_task\_completed: Whether the complete user instruction task has been fully completed.\\\\
    
    \#\# output format\\
    '''xml\\
    \textless actionReflection\textgreater [Fill in actionReflection]\textless /actionReflection\textgreater 
    \textless whether\_task\_completed\textgreater [True or False]\textless /whether\_task\_completed\textgreater 
    '''\\\\
    
    \#\# input\\
    - Initial user task instruction: [\{task\_ins\}]\\
    - Slot information: [\{slot\_info\}]\\
    - App operation knowledge: [\{app\_knowledge\}]\\
    - Subtask list: [\{subtasks\}]\\
    - Summary of historical actions: [\{action\_history\_summarization\}]\\
    - Recent action trajectory: [\{trajectory\}]\\
    - Previous action plan: [\{plan\}]\\
    - Previous action: [\{last\_action\}]\\
    - Screenshots before and after execution: (as shown in the images)\\
\bottomrule
\end{tabular}
\caption{Prompt for RA's reflection.}
\end{table*}

\begin{table*}[htbp]

\label{tab:Prompt for RA's trajectory summarization.}
\centering
\renewcommand{\arraystretch}{1.}  
\begin{tabular}{> {\arraybackslash\hspace{0.5em}}p{\dimexpr\textwidth-2\tabcolsep-2\arrayrulewidth}<{\arraybackslash\hspace{0.5em}\scriptsize}}
\toprule
\textbf{RA: Trajectory Summarization}\\
\midrule
\#\# Role Definition\\
    I am a summary agent with deep analytical capabilities, specializing in extracting key subtasks and causal chains from complex trajectories. My core abilities include:\\
    1. Analyzing key points in environment state evolution.\\
    2. Evaluating the alignment between the overall plan and the current state.\\
    3. Establishing and summarizing causal chains between subtasks.\\
    
    Important Notes:
    - Use concise titles and bullet points to summarize each subtask's main objective, causal relationships, and outcomes.\\
    - Do not list every specific action or operation.\\
    - Integrate historical summaries with recent trajectory data, extracting incremental information while avoiding repetition.\\\\
    
    \#\# Input\\
    - Historical Summary: Existing summary in the format '[\{action\_history\_summarization\}]'.\\
    - Recent Action Trajectory: Format '\{trajectory\}', where each step includes:\\
      \begin{itemize}
          \item state: Description of the current screen state (text, buttons, icons, etc.).
          \item plan: Action plan based on the current screen state.
          \item action: Specific executed operation.
          \item action\_result: Feedback or reflection on the executed operation.
      \end{itemize}\\\\
    
    \#\# Output Format\\
    '''xml\\
    \textless action\_history\_summarization\textgreater 
    Subtask Title
    Objective: Briefly describe the core goal of the subtask.
    Causal Relationship: Summarize the key causal chains and their results in this subtask.
    ...\textless /action\_history\_summarization\textgreater \\
    '''\\
    
    Notes:\\
    - Each subtask should be described only once without listing specific actions.\\
    - The summary should be an incremental update combining historical summaries with the latest trajectory data.\\
    - Keep the language concise and avoid repetition.\\\\
    
    \#\# Example Output\\
    '''xml\\
    \textless action\_history\_summarization\textgreater 
    Login Authentication
    Objective: Verify user identity to ensure secure operations.
    Causal Relationship: Successful login is a prerequisite for subsequent data synchronization.
    Data Synchronization
    Objective: Update and verify user data to maintain consistency.
    Causal Relationship: Once synchronization is completed, the app state aligns with the server state.
    \textless /action\_history\_summarization\textgreater \\
    '''\\\\
    
    \#\# Guiding Principles\\
    - Conciseness: Retain only core information without unnecessary details.\\
    - Structured Format: Use headings and bullet points to organize information.\\
    - Integration \& Refinement: Extract incremental information from historical summaries and recent trajectory data while avoiding duplication.\\
    - Highlight Key Points: Emphasize state evolution, plan-state alignment, and causal chain construction.\\\\
    
    Please strictly follow these requirements to generate the final summary, ensuring the total word count does not exceed 200 words.
    \\
\bottomrule
\end{tabular}
\caption{Prompt for RA's trajectory summarization.}
\end{table*}

\begin{table*}[htbp]

\label{tab:The first part of the prompt for DmA's decision making.}
\centering
\renewcommand{\arraystretch}{1}  
\begin{tabular}{>{\arraybackslash\hspace{0.5em}}p{\dimexpr\textwidth-2\tabcolsep-2\arrayrulewidth}<{\arraybackslash\hspace{0.5em}\scriptsize}}
\toprule
\textbf{DmA: Decision-making Part 1}\\
\midrule
\#\# Role Positioning\\
    I am the core decision-making expert for a mobile intelligent assistant with autonomous decision-making capabilities. I follow user task instructions and slot requirements, refer to mobile operation knowledge, and execute decision-making processes to operate the mobile device and complete user tasks. In specific dilemma situations, I request user input or ask the user to operate the mobile device.\\\\
    
    \#\# action\_space\\
    This module contains five action functions that simulate human interaction with a mobile device. The first row lists the selectable action functions, and the second row provides explanations. When selecting an action function, only the parameters in parentheses can be modified.\\\\
    
    \#\#\# Click(content): Click. The 'content' must include [the name of the UI element and its labeled text, as well as a detailed description of adjacent elements (above, below, left, right)]. Example: 'Click(Click the 'Select Specification' button, labeled as 'Select Spec', located to the right of the American Coffee module).' If multiple identical buttons exist on the page, additional distinguishing information is required. Note that sometimes clicking is required on a circle or box rather than text—this should be clearly specified.\\
    
    \#\#\# TYPE(text): Type. 'text' is replaced with the content entered into the input field.\\
    
    \#\#\# SLIDE(direction): Slide. 'direction' is replaced with one of: 'up', 'down', 'left', or 'right' (no punctuation allowed). 'up' means sliding from bottom to top, 'down' means sliding from top to bottom, 'left' means sliding from right to left, and 'right' means sliding from left to right.\\
    
    \#\#\# BACK(): Return to the previous page. If an incorrect page is entered or the target button is not found, choose the 'BACK' action to return and retry.\\
    
    \#\#\# HOME(): Return to the home screen.\\
    
    \#\#\# HELP(dilemma\_description): Request assistance with a dilemma. 'dilemma\_description' should provide a detailed explanation of the current dilemma, consisting of two parts: screen state and dilemma cause. If multiple selectable options exist, the dilemma cause should specify them in detail.\\
    
    \#\# Mobile Operation Requirements\\
    \{phone\_knowledge\}\\\\
    
    \#\# input\\
    - User task instruction: [\{task\_ins\}]\\
    - Reflection on the previous action: [\{actionReflection\}]\\
    - Subtask list: [\{subtasks\}]\\
    - Slot information: [\{slot\_info\}]\\
    - Recent action trajectory: \{trajectory\}\\
    - Summary of historical action trajectory: \{action\_history\_summarization\}\\
    - App usage knowledge: \{app\_knowledge\}\\
    - Mobile screen screenshot: (as shown in the image)\\\\
    
    \#\# Decision-Making Requirements\\
    - If the current screen state does not meet slot requirements, prioritize actions to adjust the state accordingly.\\
    - The chosen action must be from the action\_space.\\
    - Actions should strictly follow app usage knowledge and mobile operation requirements.\\
    - If an action is repeated three times without effect, consider modifying the action or requesting user assistance to avoid ineffective repetition.\\
    - If the expected content cannot be found after multiple attempts, use 'HELP()' to request the user to modify their request, with a clear dilemma description.\\
    - If the task requires swiping to find relevant content, reference recent action trajectories to minimize unnecessary searching.\\
    - If the page already contains pre-filled or pre-selected information that conflicts with slot information, choose an action to update it accordingly.\\
    - Decision results must not include requests for user confirmation.\\
    - Actions must fulfill all slot information requirements.\\
    - If the user changed some of the slots in the last action, you should continue to look for task-related content that matches the updated slots.\\
    - Before deciding on the 'click' action, you should confirm whether the target button is on the current page; If it doesn't exist, you can find it by swiping or other operations and then decide to 'click'\\
\bottomrule
\end{tabular}
\caption{The first part of the prompt for DmA's decision-making.}
\end{table*}

\begin{table*}[htbp]

\label{tab:The second part of the prompt for DmA's decision making.}
\centering
\renewcommand{\arraystretch}{1}  
\begin{tabular}{>{\arraybackslash\hspace{0.5em}}p{\dimexpr\textwidth-2\tabcolsep-2\arrayrulewidth}<{\arraybackslash\hspace{0.5em}\scriptsize}}
\toprule
\textbf{DmA: Decision-making Part 2}\\
\midrule
    \#\# output\\
    - state: Analyze and describe task-related elements (texts, buttons, icons, images) on the screen based on the subtask list. The output should include:
      1. A page overview. 
      2. Descriptions of relevant UI elements, including the state(such as if the input mode is activated). 
      3. Whether the page state matches user instructions and slot information.\\
    - plan: Includes the current subtask and the planned action.\\
    - action: The action chosen from the action\_space to execute the plan, formatted as per 'action\_space' syntax (e.g., 'Click(content)').\\\\
    
    \#\# output format\\
    '''xml\\
    \textless  state\textgreater [state content]\textless /state\textgreater 
    \textless plan\textgreater [plan content]\textless /plan\textgreater 
    \textless action\textgreater [action content]\textless /action\textgreater 
    '''\\\\
    
    \#\# Strict Decision-Making Process\\
    START: Perceive all screen elements, including buttons, icons, texts, and images. Analyze their functions and meanings, and determine whether the current page state fully meets slot requirements. Then, proceed to Step 2.\\
    
    Step 2: Based on task instructions, subtask list, reflection on the last action, and recent action trajectory, analyze whether task-related content (product or service) is found on the screen.\\
      - If yes, proceed to Step 2.1.\\
      - If no, proceed to Step 3.\\
    
    Step 2.1: Does the action require entering sensitive information (name, phone number, ID, password)?\\
      - If yes, decision = 'HELP(dilemma\_description)' (describe sensitive information and request user input).\\
      - If no, proceed to Step 2.2.\\
    
    Step 2.2: Does the action involve login, verification, payment, order submission, confirmation, or repeated failed attempts?\\
      - If yes, decision = 'HELP(dilemma\_description)' (describe sensitive action and request user input).\\
      - If no, proceed to Step 2.3.\\
    
    Step 2.3: Does task-related information on the page match user instructions and slot information?\\
      - If yes, proceed to Step 2.3.1.\\
      - If no, proceed to Step 2.3.2.\\
    
    Step 2.3.1: Can conflicting information be changed via click or input?\\
      - If yes, proceed to Step 2.3.2.\\
      - If no, decision = 'HELP(dilemma\_description)' (describe conflicting information).\\
    
    Step 2.3.2: Execute actions based on user instructions and supplementary information, then END.\\
    
    Step 2.4: Execute actions on relevant screen elements, then END.\\
    
    Step 3: Has task-related content been searched twice in history?\\
      - If yes, decision = 'HELP(dilemma\_description)' (describe missing content).\\
      - If no, proceed to Step 4.\\
    
    Step 4: Continue searching for task-related content, then END.\\
    
    END\\
\bottomrule
\end{tabular}
\caption{The second part of the prompt for DmA's decision-making.}
\end{table*}

\begin{table*}[htbp]

\label{tab:Prompt for DmA's task decomposition.}
\centering
\renewcommand{\arraystretch}{1.4}  
\begin{tabular}{>{\hspace{0.5em}}p{\dimexpr\textwidth-2\tabcolsep-2\arrayrulewidth}<{\hspace{0.5em}}}
\toprule
\textbf{DmA: Task Decomposition}\\
\midrule
\#\# Role and Task\\
    I am a task decomposition and planning expert for mobile operations. My task is to generate a structured task plan based on the given task instruction and existing device operation knowledge by breaking down the initial task into multiple subtasks.\\\\
    
    \#\# Planning Requirements\\
    - Subtasks must be concise and should not include overly detailed task-specific information (e.g., specific product names or preference settings). They should be arranged in a logical sequence.\\
    - Incorrect subtask: 'Select Less Ice option' (contains specific details like 'Less Ice')\\
    - Correct subtask: 'Select product specifications'\\
    - Subtasks should not include information confirmation steps.\\
    - The first step is usually to open the corresponding app based on the task intent. It is assumed that the relevant app is already installed, so there is no need to include app installation steps.\\
    - If a subtask involves login, verification, or payment, it must explicitly request user action and include the phrase 'Request User'.\\
    - The subtask order and format must strictly follow app operation knowledge.\\\\
    
    \#\# Input\\
    - User Task Instruction: [\{task\_ins\}]\\
    - App Operation Knowledge: [\{app\_knowledge\}]\\\\
    
    \#\# Output Format\\
    \textless subtask\textgreater [Fill in subtask1]\textless /subtask\textgreater \textless subtask\textgreater [Fill in subtask2]\textless /subtask\textgreater ...\\
\bottomrule
\end{tabular}
\caption{Prompt for DmA's task decomposition.}
\end{table*}

\end{document}